\documentclass{article}
\usepackage{ijcai24}

\usepackage{times}
\usepackage{soul}
\usepackage{url}
\usepackage[hidelinks]{hyperref}
\usepackage[utf8]{inputenc}
\usepackage[small]{caption}
\usepackage{graphicx}
\usepackage{amsmath}
\usepackage{amsthm}
\usepackage{booktabs}
\usepackage{algorithm}
\usepackage{algorithmic}
\usepackage{multirow} 
\usepackage{multicol} 
\usepackage{arydshln} 
\usepackage{amssymb} 
\usepackage{subfigure} 
\usepackage{subcaption} 
\usepackage{color} 
\usepackage{enumitem} 
\usepackage[switch]{lineno}
\title{From Pixels to Progress: Generating Road Network from Satellite Imagery for Socioeconomic Insights in Impoverished Areas}

\author{
Yanxin Xi$^1$
\and
Yu Liu$^2$
\and
Zhicheng Liu$^2$
\and
Sasu Tarkoma$^1$
\and
Pan Hui$^{1,3}$
\and
Yong Li$^2$\\
\affiliations
$^1$University of Helsinki, Department of Computer Science\\
$^2$Tsinghua University, Department of Electronic Engineering, Beijing National Research Center for Information Science and Technology (BNRist)\\
$^3$Hong Kong University of Science and Technology (Guangzhou), Computational Media and Arts Thrust\\
\emails
liuyu2419@126.com,
liyong07@tsinghua.edu.cn
}

\begin{document}

\maketitle

\begin{abstract}
The Sustainable Development Goals (SDGs) aim to resolve societal challenges, such as eradicating poverty and improving the lives of vulnerable populations in impoverished areas. Those areas rely on road infrastructure construction to promote accessibility and economic development. 
Although publicly available data like  OpenStreetMap is available to monitor road status, data completeness in impoverished areas is limited. Meanwhile, the development of deep learning techniques and satellite imagery shows excellent potential for earth monitoring. To tackle the challenge of road network assessment in impoverished areas, we develop a systematic road extraction framework combining an encoder-decoder architecture and morphological operations on satellite imagery, offering an integrated workflow for interdisciplinary researchers.
Extensive experiments of road network extraction on real-world data in impoverished regions achieve a $42.7\%$ enhancement in the F1-score over the baseline methods and reconstruct about $80\%$ of the actual roads. We also propose a comprehensive road network dataset covering approximately $794$,$178$ $km^2$ area and $17.048$ million people in $382$ impoverished counties in China.
The generated dataset is further utilized to conduct socioeconomic analysis in impoverished counties, showing that road network construction positively impacts regional economic development. The technical appendix, code, and generated dataset can be found at \href{https://github.com/tsinghua-fib-lab/Road_network_extraction_impoverished_counties}{https://github.com/tsinghua-fib-lab/Road\_network\_extraction\_impoverished\_counties}.
\end{abstract}

\maketitle

\section{Introduction}

Vulnerable populations in impoverished regions (e.g., poor areas distant from cities) encompass groups experiencing risks due to socioeconomic, health, or environmental factors \cite{wang2023incentive}. Prioritizing resolving the societal challenges for vulnerable populations is crucial for achieving Sustainable Development Goals (SDGs) \cite{Transforming2015,Nath2016making} and Leave No One Behind (LNOB) \cite{LNOB}. 
As a Chinese saying goes, ``If you want to be wealthy, build roads first.'', to improve the living quality for people and boost the economic development in impoverished regions, the transportation infrastructure development, which facilitates internal transportation and attracts external investments, has been proved effective \cite{sebayang2020infrastructure,zhang2021infrastructure,banerjee2020road}. Therefore, timely and accurate monitoring of road network development is essential for reducing the population left behind because of the vulnerabilities. 

\begin{figure}[tb]
  \centering
  \subfigure[Satellite image]{
    \centering
    \includegraphics[width=0.3\linewidth]{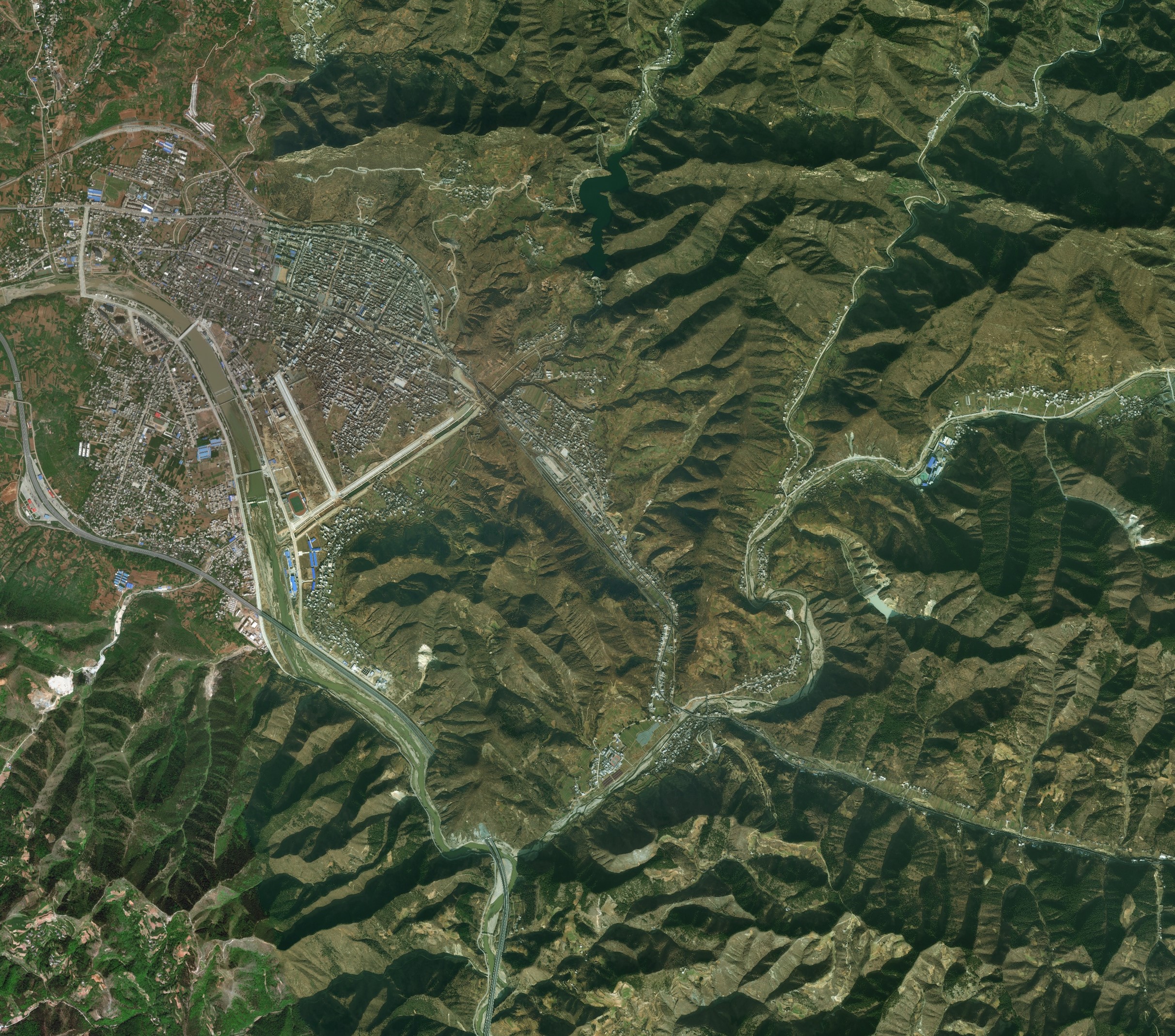}
}
    \subfigure[OSM roads]{
    \centering
    \includegraphics[width=0.3\linewidth]{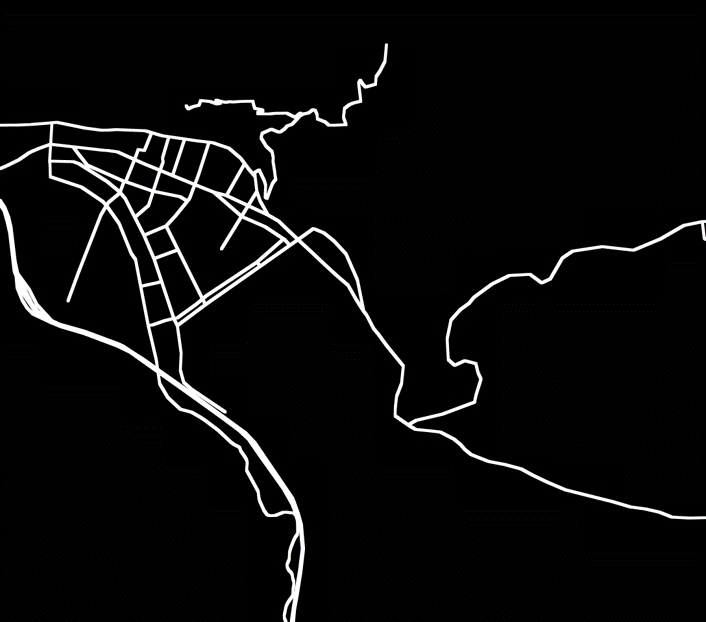}
}
    \subfigure[Generated roads]{
    \centering
    \includegraphics[width=0.3\linewidth]{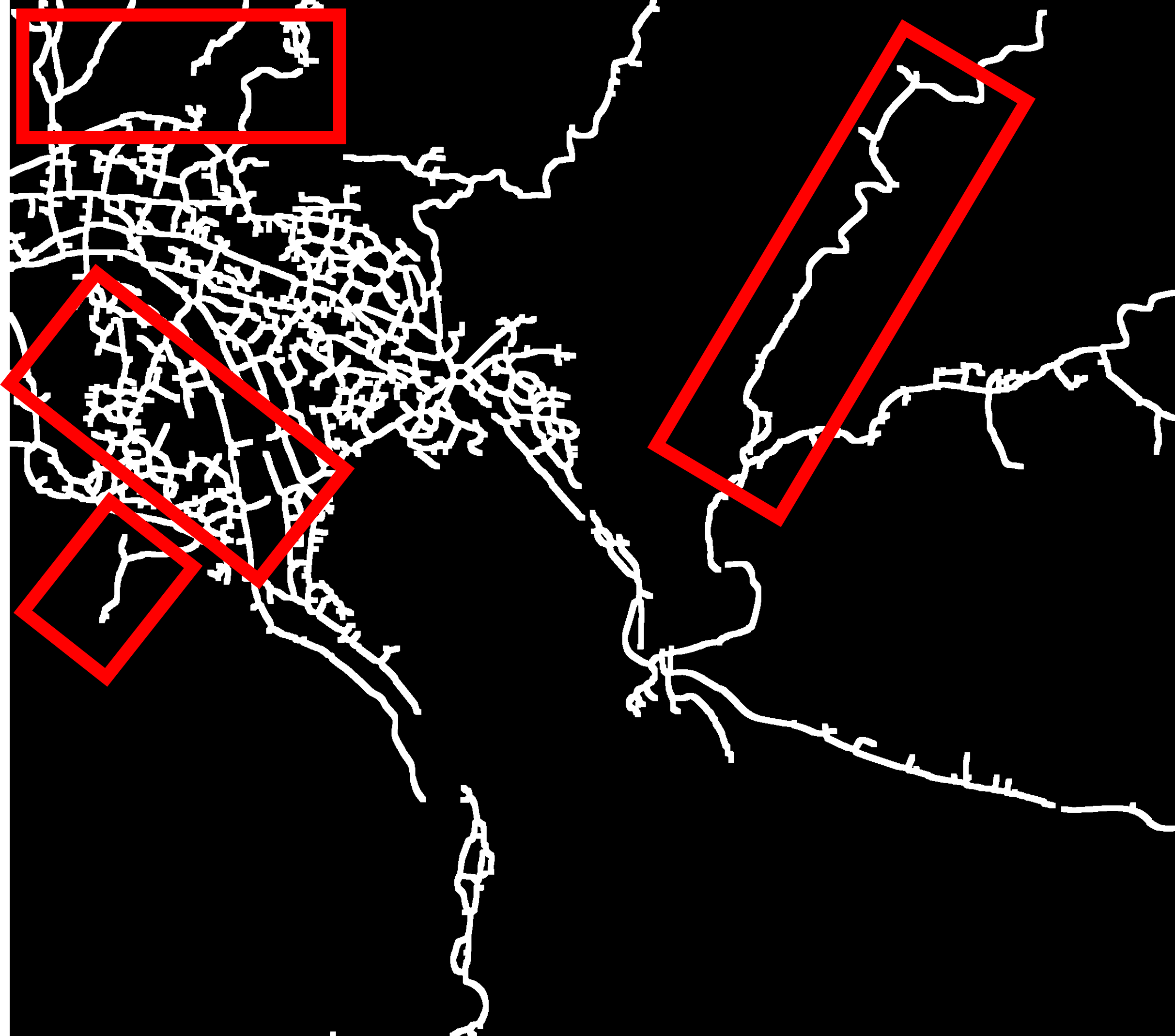}
}
  \caption{Comparison of (a) one satellite image, (b) corresponding OSM road network, and (c) our generated road network. (Latitude=$33.6945^\circ$N,  Longitude=$110.3237^\circ$E). The roads in red rectangles in figure (c) are missing in OSM data.}
  \label{fig:sat-OSM}
\end{figure}


However, the current road network data in impoverished regions has several limitations. First, the commonly used OpenStreetMap (OSM) data exhibits significant deficiencies. OSM is a volunteered geographic information platform whose completeness relies on the annotations from volunteer contributors \cite{haklay2008openstreetmap,vargas2020openstreetmap,zhou2019investigating}. For example, Figure \ref{fig:sat-OSM} shows an area in Danfeng County in Shaanxi Province, China, and its corresponding OSM and our produced road network. We can see that within the populated areas, some short connecting roads and certain major roads that link town centers of external areas through mountains are missing in OSM data. Second, the road data from official agencies covers roughly coarse-level survey data, possesses a time lag, and merely provides road length information, which prohibits interdisciplinary data users from accessing real-time and detailed road data. At last, purchasing road data from map platform corporations is unaffordable for most researchers and organizations. Those shortcomings hinder real-time road monitoring, especially in impoverished counties.

Meanwhile, various deep learning approaches for segmentation based on satellite imagery have recently been proposed \cite{yang2022road,mei2021coanet,zhang2018road,zhang2024uv}. Combined with the development of large-scale road annotation datasets \cite{van2018spacenet,mnih2013machine,wang2021loveda}, the segmentation approaches recognize roads from satellite imagery. Approved by the advantages of long temporal spanning, comprehensive spatial coverage, and high spatial resolution, the widely used satellite imagery is a better data source for observing roads in impoverished regions with high accuracy \cite{mei2021coanet,xi2023satellite,Nachmany2019}. 

We propose a systematic framework for extracting road networks in impoverished regions from satellite imagery. Specifically, the framework consists of the data pre-processing, road segmentation, image morphological operations, and graph extraction steps. (1) The data pre-processing step adopts the Laplacian variance \cite{bansal2016blur} and pixel intensity \cite{he2010single} metric to filter out the invalid data in the satellite images. (2) Considering the occlusion in roads in satellite imagery, we use an encoder-decoder-based road extraction model including strip convolution and connectivity attention in the road segmentation step. (3) The image morphological operations include the morphological closing operation and skeletonization to transform the road mask into the road centerline binary image. (4) A combustion algorithm is applied in the graph extraction step to generate the road network graph with nodes and edges from the road centerline image. We evaluate our designed framework on satellite imagery and real-world road network data in impoverished counties in China. Besides, we propose a comprehensive road network dataset in $382$ impoverished counties.

The main contributions of this work are summarized as follows:
\begin{itemize}
     \item We propose a scalable framework for extracting county-level road networks from satellite imagery in impoverished counties. We introduce statistical methods to filter irregular satellite imagery, and leverage an encoder-decoder architecture preserving road connectivity to segment road masks from satellite imagery. Through morphological operations and combustion algorithm, the county-level road network data is generated.
     \item To the best of our knowledge, we are the first to produce a fine-grained road network dataset covering $17.048$ million people in $382$ impoverished counties in China. The dataset will be beneficial for further SDG research about road infrastructure.
     \item Extensive experiments on real-world road data generation in impoverished counties show that our proposed framework achieves a $42.7\%$ increase in F1-score compared with baselines and road length reconstruction rate of about $80\%$. We also conduct socioeconomic analysis on our produced road network dataset. The analysis shows that GDP in impoverished counties can be increased by $7\%$ unit quantitatively through road expansion.
\end{itemize}
%
\section{Related Work}
\subsection{Satellite Imagery-based Road Extraction}
With the development of deep learning techniques, satellite imagery has been utilized for road extraction. \cite{mnih2010learning} utilize the convolutional neural network (CNN) with encoder-decoder architecture to detect roads in high-resolution satellite images. \cite{zhang2018road} apply a deep residual U-Net to segment roads to optimize the training procedure.
To overcome the road occlusions caused by surroundings, \cite{yang2022road} introduce RCFSNet with a multiscale context extraction module and a full-stage feature fusion module to enhance the road area's segmentation quality. To guarantee the connectivity of roads, \cite{wegner2015road} adopt the shortest path method as a post-processing step to improve the segmentation results. \cite{liu2018roadnet} further combine the edges and centerlines of roads for better-connected segmentation. \cite{mei2021coanet} propose a skip convolution module and connectivity attention module to guarantee road connectivity further and remove occlusion effects. Those works only consider extracting road masks in limited satellite images and neglect to extract the large-scale road network in impoverished areas for practical SDG analysis. 


\subsection{Transportation Infrastructure-based Socioeconomic Study}
Many scholars have explored the effect of transportation infrastructure on economic development on large scales like country or state. \cite{ighodaro2009transport} studies the economic impact of transportation at the country level in Nigeria.
\cite{pradhan2013effect} delve into the impact of transportation infrastructure on India's economic growth and reveal a unidirectional causality from rail transportation to economic growth. 
\cite{Zhang2023role1} study the relationship between transportation infrastructure development and economic growth in the UK. \cite{banerjee2020road} assess the impact of access to transportation networks on regional demographic and economic outcomes in China. At a smaller scale, such as the city, the transportation network, especially the road network's relationship with socioeconomic activities, is studied. \cite{porta2009street} study the correlation between road network centrality and economic activities. \cite{spadon2018topological} utilize the road network topological features to predict demography. \cite{zhou2022road} develop a theoretical framework using panel data to illustrate how road construction facilitates economic growth and poverty alleviation. However, research on road network's socioeconomic impact in impoverished areas is still scarce. 

Our work offers a road network generation framework and corresponding generated dataset for this missing research area, providing relevant researchers and policymakers further help in supporting impoverished counties. 
\section{Method}
\begin{figure*}[t]
    \centering
    \includegraphics[width=0.99\linewidth]{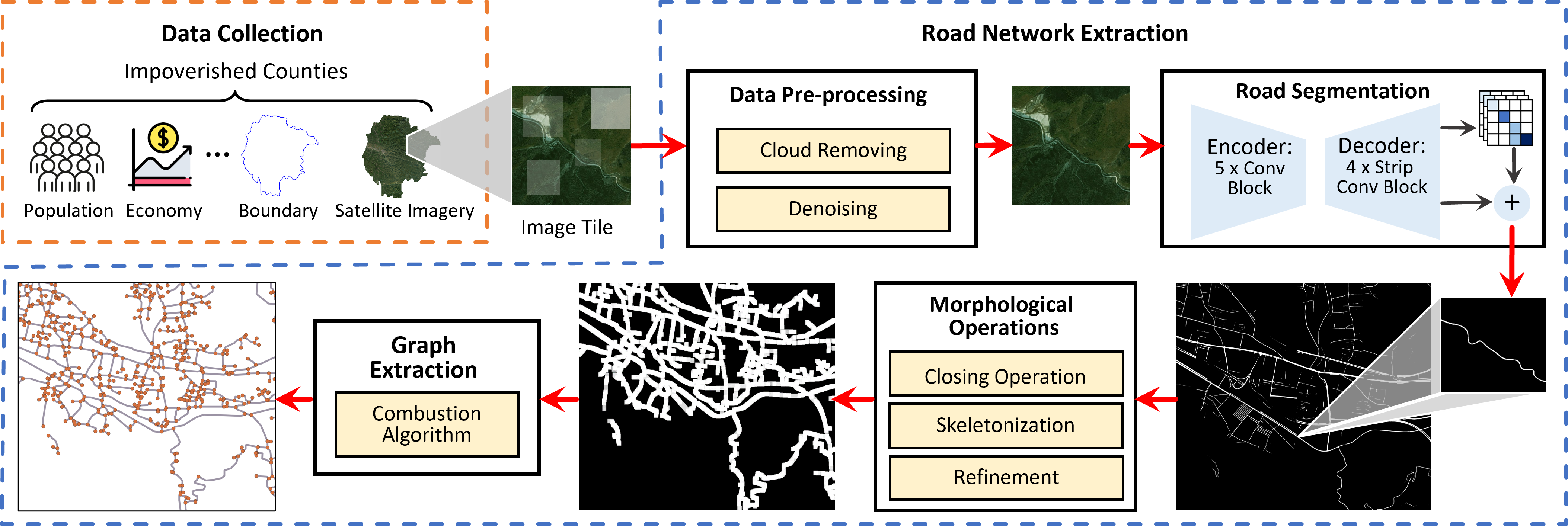}
    \caption{The overall framework of the proposed road network extraction framework.}
    \label{fig:enter-label}
\end{figure*}
In this section, \textbf{we transform the task of road network generation into a road segmentation task based on satellite imagery in our framework}. The overall framework is shown in Figure \ref{fig:enter-label}. 
We first introduce statistical methods to filter the satellite image tiles, primarily to remove certain noisy or cloud-covered image tiles. Secondly, we apply a road segmentation model that jointly considers the segmentation of roads and the relationship between neighboring road pixels to generate binary road masks from satellite image tiles. Thirdly, we refine and convert the binary road masks into an ``one-pixel width'' road centerline image and finally extract the road network using a combustion algorithm.

\subsection{Data Pre-processing}
This step aims to reduce the effect of invalid image tiles on the road segmentation results. Due to weather conditions or transmission signal disruptions, there will be cloud-covered or noisy satellite images. All those images introduce errors in the road segmentation results and further affect the socioeconomic analysis in impoverished regions. Therefore, to identify the noisy or cloud-covered images, we apply the Laplacian variance and pixel intensity metric, respectively.

Specifically, we apply the Laplacian variance \cite{bansal2016blur} value to mark the noisy satellite image tiles. Laplacian variance measures the amount of high-frequency detail in an image. Given an RGB (Red, Green, Blue) image $I(x, y, z)$ where $x$ and $y$ mark the pixel coordinates of an image, and $z$ indicates the channel (R, G, B), the Laplacian image $L(x, y, z)$ for channel $z$ is calculated as:
\begin{equation}
    L(x,y,z) = \nabla^{2}I(x,y,z) = \frac{\partial^2I}{\partial^2x} + \frac{\partial^2I}{\partial^2y},
\end{equation}
where $\frac{\partial^2I}{\partial^2x}$ and $\frac{\partial^2I}{\partial^2y}$ represent the second derivatives for the $x$ and $y$ coordinates in the grayscale image of channel $z$. The overall Laplacian variance for an RGB image $I$ is calculated as: 
\begin{gather}
     Var_{\mathrm{Lap}} = \frac{\sum_{z \in \{R, G, B\}} \mathbb{E}[{(L(x, y,z) - \mu^z)}^2]}{3},
\end{gather}
where $\mathbb{E}$ denotes the expectation operator
and $\mu^z$ is the mean pixel value of $L(:,:,z)$. By setting a threshold $Var^{T}_{\mathrm{Lap}}$, the images that satisfy $Var_{\mathrm{Lap}} >  Var^{T}_{\mathrm{Lap}}$ are identified as images with high-frequency noise.

For the cloud-covered image tiles, we follow the image prior that normal images contain some pixels with very low intensities (pixel values) in at least one color channel \cite{he2010single}. 
For an RGB image $I(x, y, z)$, the mean normalized pixel intensity \cite{gonzales1987digital} for channel $z$ is calculated as 
\begin{equation}
    Mean^z_{\mathrm{Int}} = \mathbb{E}[\frac{I(:, :, z)}{255}].
\end{equation}
And the images with $Mean^z_{\mathrm{Int}}$ for any channel $z$ ($z\in \{R,G,B\}$) below a threshold $Mean^T_{\mathrm{Int}}$ are selected as the non-cloud image tiles. 

Until this step, we determine the regular images for the following road segmentation task. For the cloud-covered or noisy data, we apply temporal interpolation to guarantee the completeness of the dataset.


\subsection{Road Segmentation}
This step generates the road masks from satellite images by developing an encoder-decoder-based road extraction model. Specifically, we adopt the connectivity attention network (CoANet) \cite{mei2021coanet}, shown in Figure 2, to extract roads from satellite image tiles. CoANet is composed of an encoder-decoder architecture, where the encoder has five convolutional blocks for feature learning. 
The decoder contains four strip convolution blocks composed of strip convolutions with horizontal, vertical, left diagonal, and right diagonal directions that fit the shape of the roads and extract the linear features. The segmentation result from the decoder and the connectivity cube are combined to generate the final road mask with better connectivity. 


Given a satellite image $I$ as input, the CoANet model generates the binary road mask output $S$,
\begin{equation}
    S=\mathrm{CoANet}(I).
\end{equation}

\subsection{Morphological Operations}
We perform post-processing on the road masks to further generate the road centerlines. Specifically, we sequentially apply the morphological closing operation, skeletonization, and refinement on the road masks. 
\begin{itemize}[leftmargin=*]
    \item \textbf{Morphological Closing Operation} \cite{gonzales1987digital}: The binary road mask may contain holes because of the occlusions of trees and shadows. Therefore, we apply the morphological closing operation, which first dilates the binary image and then erodes the dilated image with the same kernel $K_{\mathrm{morph}}$. Given a binary road segmentation image $S$, the closing operation is computed as
    \begin{equation}
    S \bullet K_{\mathrm{morph}} = (S\oplus K_{\mathrm{morph}}) \ominus K_{\mathrm{morph}},
    \end{equation}
    where $\oplus$ is dilation and $\ominus$ is erosion operation. It fills the small holes and preserves the shape and size of road masks.

    \item \textbf{Skeletonization} \cite{zhang1984fast}: This step extracts the road centerline from the binary road mask. Since the roads take several pixels in width in the satellite images, the road masks contain roads of the same width. For a binary image, skeletonization is to identify and remove the foreground border
    pixels and thin down the foreground to a ``skeleton" of unitary thickness by reserving the connectivity.  
  
    \item \textbf{Refinement}: Roads are interconnected objects connecting different regions' locations. Therefore, in this step, we remove the isolated foreground pixels, which might be the misclassified pixels in the uninhabited areas. To achieve this goal, we first determine all the connected components with $4$-connectivity \cite{gonzales1987digital} in the road skeleton image and remove the connected components below a certain length threshold $\mathrm{T}_{\mathrm{refine}}$. 
  
\end{itemize}

\subsection{Graph Extraction}
This step extracts the road network graph with node and edge structures from the refined road centerline binary image. We apply the commonly used combustion algorithm \cite{shi2009automatic,ruan2020learning} to construct an undirected graph. It first detects crossing pixels on the road centerline binary image and adds nodes on the road skeleton. Specifically, for each pixel, the number of road fringe $n$ (i.e., how many road segments connect to this pixel) in a neighboring area is first calculated, and those pixels with $n = 1$ (road endpoint) or $n > 2$ (more than two road segments intersect) are considered crossing pixels. Then, the neighboring crossing pixel along each road segment is located for each crossing pixel. The road pixels are added as nodes at a regular interval between each pair of crossing pixels. Finally, the image coordinates of the node pixels are transformed into geographic coordinates, and an edge is built between two neighboring nodes connected by road skeleton.
\section{Experiments}
\subsection{Experimental Setup}

\subsubsection{Satellite Imagery}
We use the RGB satellite imagery acquired from the Environmental Systems Research Institute (Esri)\footnote{\href{https://www.esri.com/en-us/home}{https://www.esri.com/en-us/home.}}. We select the satellite image tiles of zoom level $17$. And the spatial resolution is about $1.2$ $m$, where roads can be recognized manually. 
Considering the image quality and temporal need for studying the road network variation, we select satellite image tiles in 2017 and 2021. 

\subsubsection{Road Network Ground Truth}
The ground truth data is provided by a map platform corporation in China.  
It provides the geographic coordinates of each road segment, road id, and road type. We evaluate the road network extraction results in $2017$ and $2021$, respectively.

\subsubsection{Implementation Details}
We randomly select $20$ cloud-covered, noisy, and regular satellite image tiles, respectively, in data pre-processing step. Then, based on the selected samples, we set the thresholds $Var^{T}_{\mathrm{Lap}}=10000$ and $Mean^T_{\mathrm{Int}} = 0.45$ empirically. The CoANet model pre-trained on the DeepGlobe road segmentation dataset with default hyperparameters is selected \cite{CoANetGit}. 
In the morphological operations, the morphological kernel for closing operation $K_{\mathrm{morph}} = 11$ and the threshold $\mathrm{T}_{\mathrm{refine}} = 500$ are determined by experiment trials. For validating the road network extraction performance, we randomly select 10 impoverished counties, of which the statistics are in A.1. The experiments are conducted on an NVIDIA GeForce RTX 2080 Ti GPU (11GB) with 126GB RAM. 

\begin{table*}[htb]
    \centering
     \resizebox{1\textwidth}{!}{
    \begin{tabular}{c|cccccc|cccccc}
    \hline
    Year & \multicolumn{6}{|c|}{2017}& \multicolumn{6}{|c}{2021} \\
    \hline
    Model&Precision$\uparrow$ &Recall$\uparrow$ &F1-score$\uparrow$  &MRL $\downarrow $& MRD $\downarrow$ & RI\texttt{@}3$\uparrow$&Precision$\uparrow$&Recall$\uparrow$&F1-score$\uparrow$&MRL$\downarrow$ &MRD$\downarrow$ &RI\texttt{@}3$\uparrow$\\
    \hline
    RCFSNet &0.8504&0.3605&0.4358&0.7937&16.487&0.0676&\textbf{0.9538}&0.2015&0.3215&0.9078&48.467&0.0421\\		
    ViT &0.6980&0.5880&0.5752&0.8522&14.411&0.1293&0.7727&0.4621&0.5350&0.9310&20.099&0.0572\\	
    OSM &\textbf{0.9633}&0.3168&0.4575&0.6716&2.7578&0.1734&0.9513&0.4237&0.5621&0.5246&1.6900&0.2173\\
    \hline
    Ours &0.7929&\textbf{0.7313}&\textbf{0.7514}&\textbf{0.2117}&\textbf{0.3602}&\textbf{0.6616}&0.8125&\textbf{0.7316}&\textbf{0.7638}&\textbf{0.1457}&\textbf{0.4709}&\textbf{0.6049}\\	
    \hline
   \end{tabular}
   }
   \caption{Road network extraction results in $10$ counties in 2017 and 2021, respectively. The best results are in bold.}
    \label{tab:my_label2}
\end{table*}

    


\subsubsection{Evaluation Metrics}
The accuracy of a road network generated from satellite imagery relies on geometry and topology. Therefore, following the methods in the literature \cite{li2022generating,ruan2020learning}, we apply the commonly used metrics: precision, recall, and F$1$-score in the graph sampling method \cite{biagioni2012inferring,aguilar2021graph} for jointly evaluating the geometric and topological accuracy of the extracted road networks. To improve computational efficiency, we simplify curved road segments into straight lines. Besides, we also compare the reconstruction rate of road intersection connected to $3$ or more road segments (RI@$3$), mean absolute percentage error of the road length (MRL) and road network density (MRD) between the generated and ground truth road network. The computations are presented in A.2.

\subsubsection{Baselines}
To evaluate the effectiveness of our road network extraction framework, we compare it with three baselines. 
\begin{itemize}[leftmargin=*]
    \item RCFSNet \cite{yang2022road}. This method adopts an encoder-decoder network that extracts road context and integrates full-stage features for road extraction from satellite imagery. 
    The model is trained on the Massachusetts road dataset \cite{mnih2013machine}.
    \item ViT \cite{chen2022vitadapter}. Vision Transformer-Adapter-based semantic segmentation model. The backbone is a plain vision transformer that learns representations from large-scale multi-modal data. 
    The model is trained on the LoveDA dataset \cite{wang2021loveda} for road segmentation.
    \item OSM \cite{OSM2022}. 
    We crop the OSM road network from the GeoFabrik historical files according to the selected counties' boundaries \cite{GaryBikini/ChinaAdminDivisonSHP} and years.
\end{itemize}

\begin{figure*}[tb]
  \centering
  \subfigure[RCFSNet (county 1)]{
    \centering
    \includegraphics[width=0.23\linewidth]{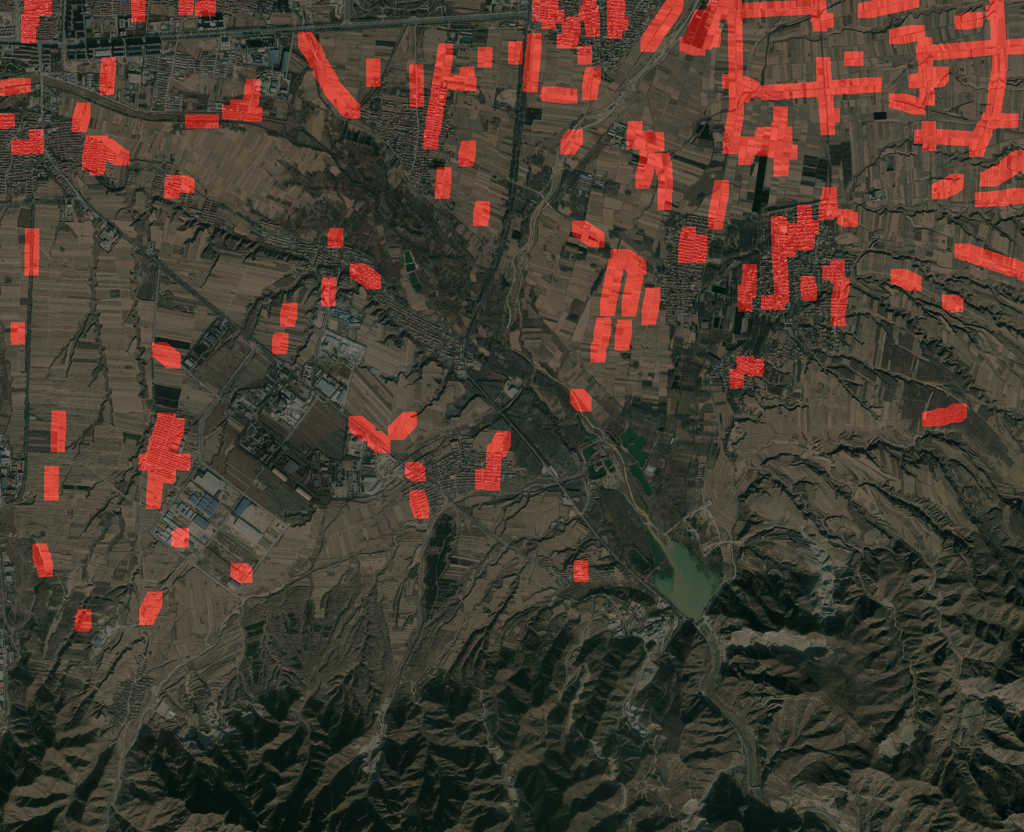}
    \label{fig:sub1}}
  \subfigure[ViT (county 1)]{
    \centering
    \includegraphics[width=0.23\linewidth]{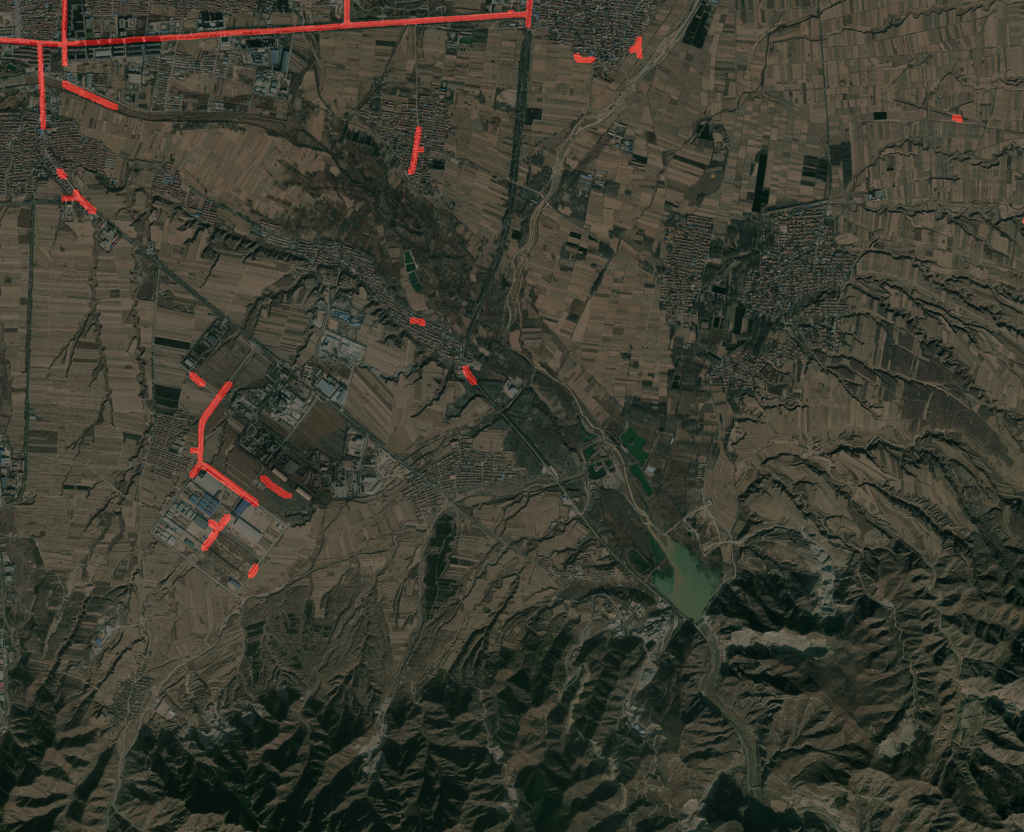}
    \label{fig:sub2}
}
    \subfigure[OSM (county 1)]{
    \centering
    \includegraphics[width=0.23\linewidth]{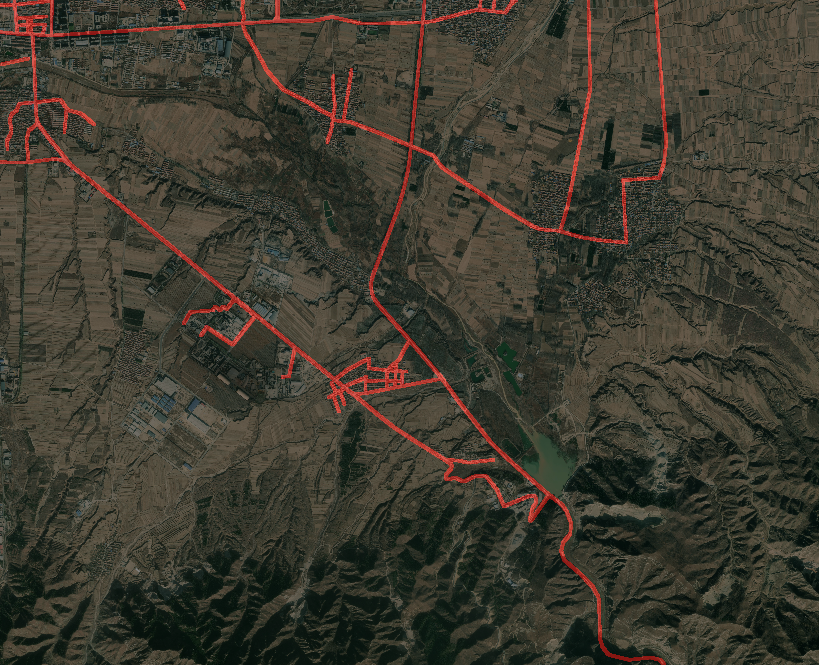}
    \label{fig:sub3}}
    \subfigure[Ours (county 1)]{
    \centering
    \includegraphics[width=0.23\linewidth]{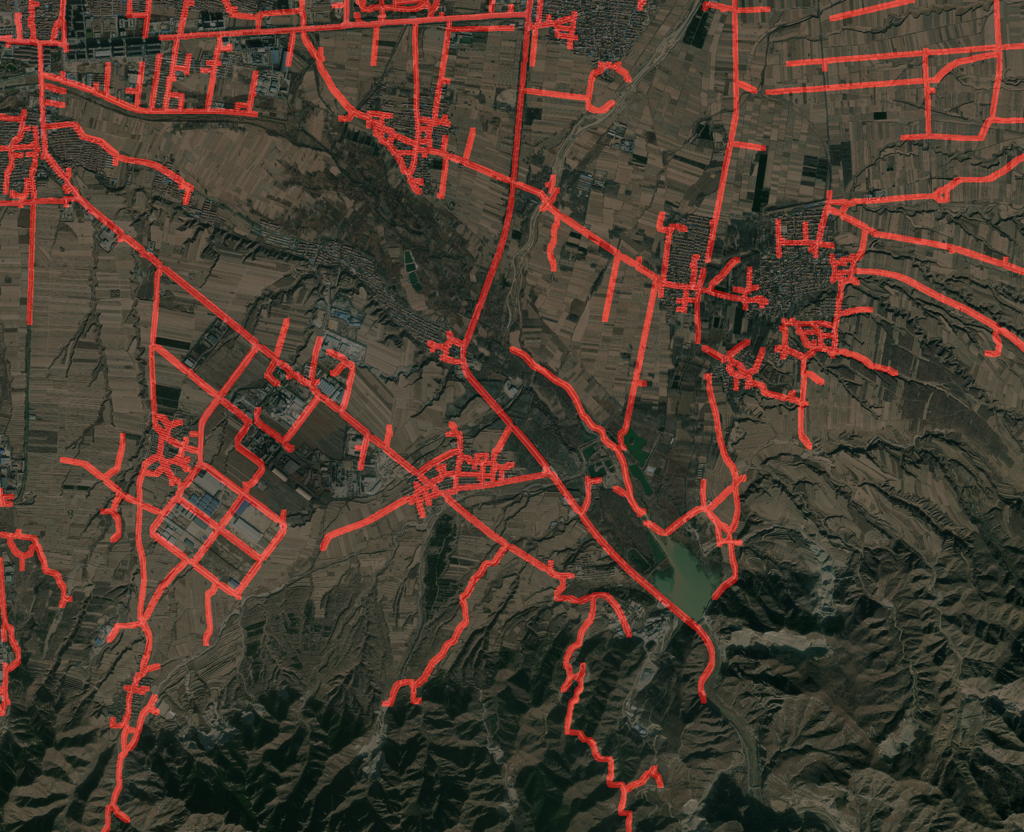}
    \label{fig:sub4}
}
\\
  \subfigure[RCFSNet (county 2)]{
    \centering
    \includegraphics[width=0.23\linewidth]{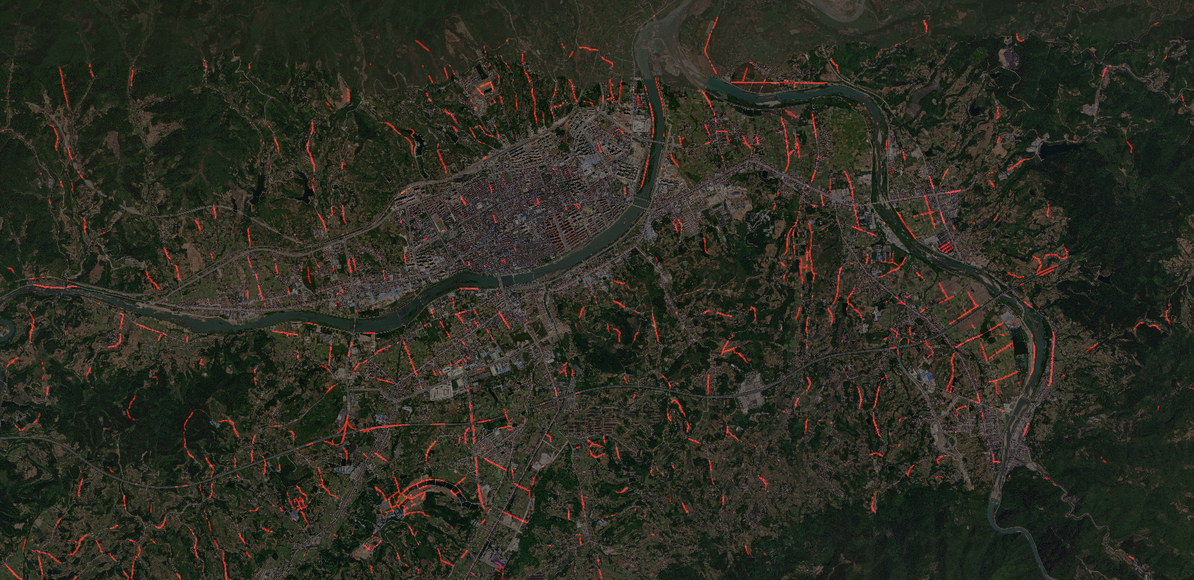}
    \label{fig:sub5}
}
  \subfigure[ViT (county 2)]{
    \centering
    \includegraphics[width=0.23\linewidth]{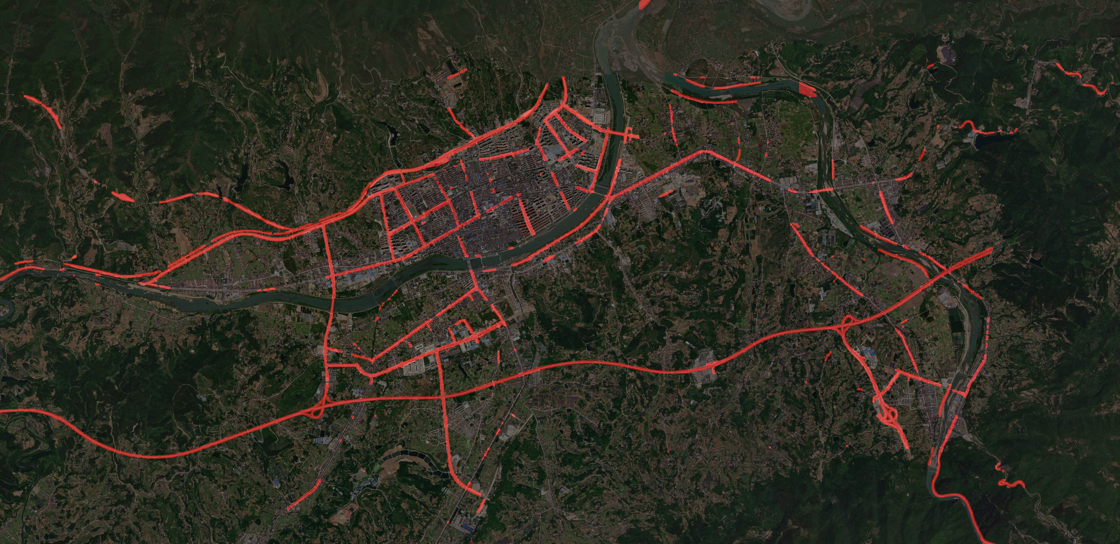}
    \label{fig:sub6}
}
    \subfigure[OSM (county 2)]{
    \centering
    \includegraphics[width=0.23\linewidth]{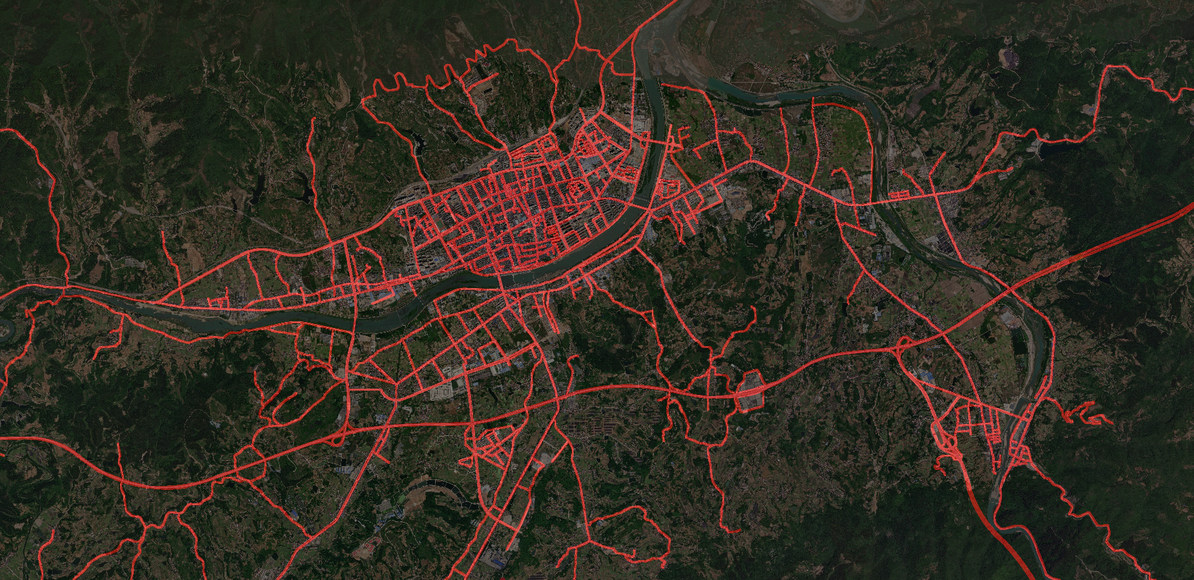}
    \label{fig:sub7}
}
    \subfigure[Ours (county 2)]{
    \centering
    \includegraphics[width=0.23\linewidth]{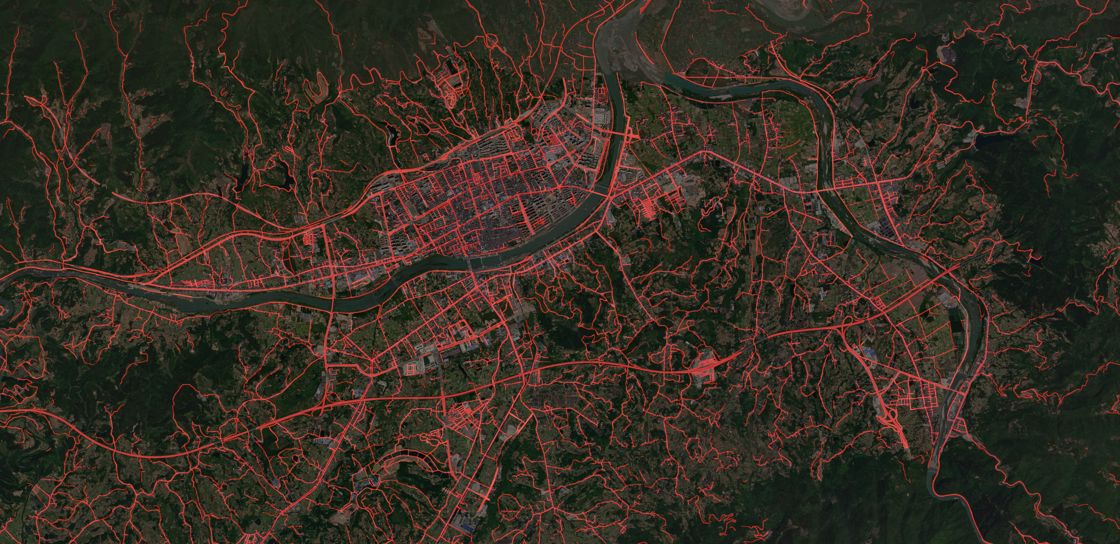}
    \label{fig:sub8}
}
  \caption{Visualization of the road network extraction results in two impoverished counties. The roads are plotted in red. The locations of the two counties are (Latitude=$39.4255^\circ$N,  Longitude=$114.2517^\circ$E) and (Latitude=$32.9858^\circ$N,  Longitude=$107.7386^\circ$E).}
  \label{fig:qualitiveVis}
\end{figure*}

\subsection{Performance Comparison}
We report the average evaluation metrics on $10$ randomly selected impoverished counties in $2017$ and $2021$ in Table \ref{tab:my_label2}. Overall, our proposed framework shows the best performance on all evaluation metrics except precision, showing that our framework is effective in county-scale road network generation from satellite imagery. Besides, our generated dataset shows a much higher recall and F1-score than the baselines, which reveals that from the perspective of topology and geometry, more real roads in impoverished counties can be recovered by our method. OSM showed high precision in both years because the roads in OSM are generated by volunteers who focus on the main roads and might have reference information when annotating the roads. ViT shows a second high recall and F1-score, which is a possible explanation for the strong performance of the vision transformer on the segmentation task. RCFSNet has a higher precision than our method. However, the recall is much lower, which may be attributed to the relatively limited transferability of a model trained on the Massachusetts dataset. Regarding road length and intersection reconstruction metrics, our framework shows the best results, which supports the effectiveness of our produced road network dataset for $382$ impoverished counties. 


Then, we qualitatively evaluate the effectiveness of our generated dataset. Specifically, we select two areas from two impoverished counties (Lingqiu County and Xixiang County) with different populations. We overlap the extracted road network on the corresponding satellite imagery shown in Figure \ref{fig:qualitiveVis}. In county 1, RCFSNet misclassifies the mountainous areas as roads. Vit can identify the roads, but the connectivity of roads is row.
OSM shows the main roads in the mountainous areas, but it omits the other roads that lead to the industrial area or connect the households in a village. Our generated dataset covers most roads, including the primary and pavement roads. In county 2, RCFSNet shows discrete roads. 
ViT and OSM capture most main roads, such as the roads in the county center. However, our generated dataset shows the roads that go into mountainous areas and connect the neighborhoods. In conclusion, our proposed dataset covers more roads than baselines and guarantees the connectivity of roads.

\subsection{Case Study}

\begin{figure}[t]
    \centering
    \includegraphics[width=0.8\linewidth]{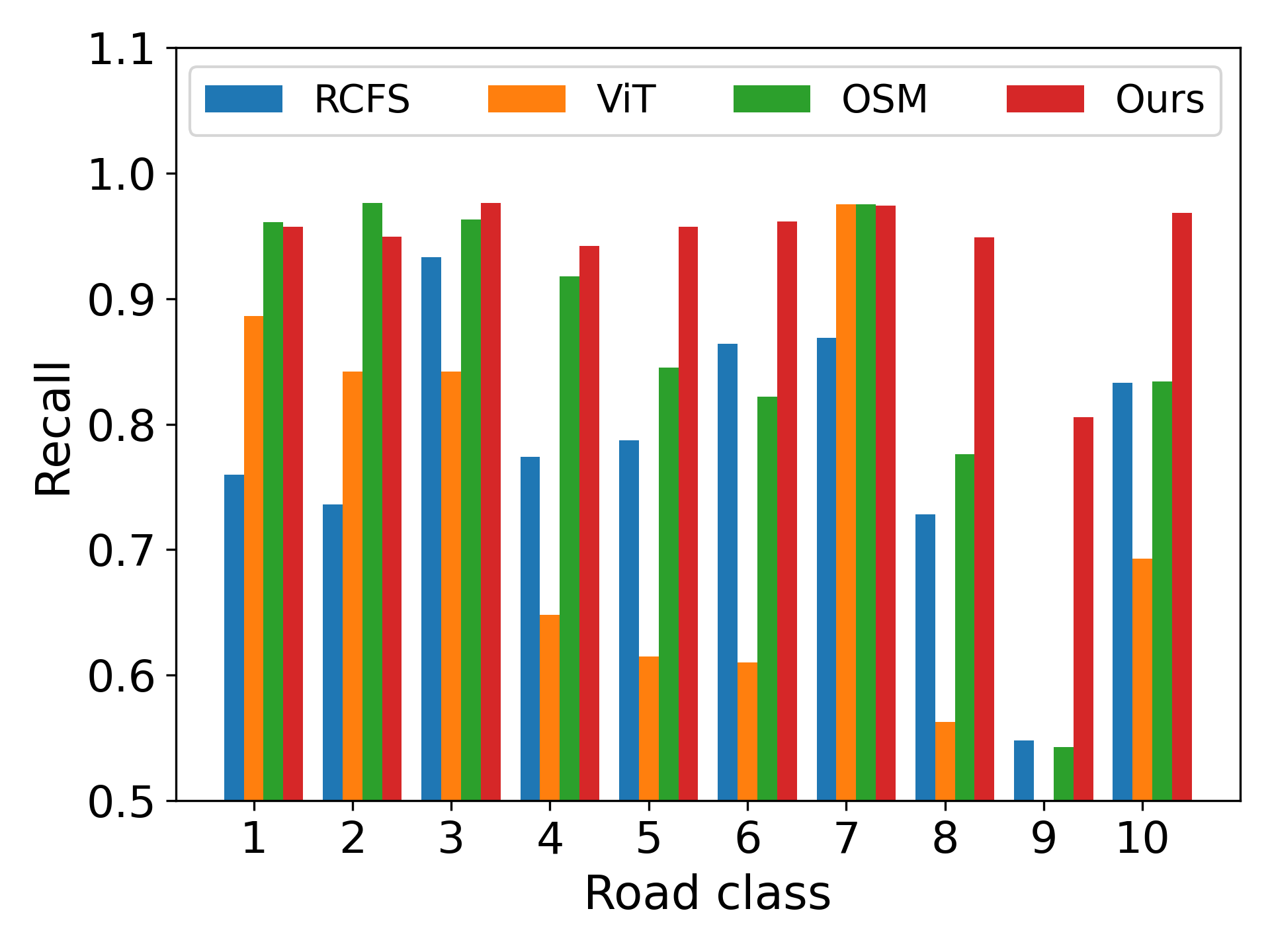}
    \caption{The results of reconstructing different classes of roads in $2021$.}
    \label{ablation}
\end{figure}

Next, we investigate the ability of our proposed framework to reconstruct different types of roads. Based on the width and speed limit of the roads, the ground truth road network is categorized into ten classes, i.e., class 1 being the highest- and class 10 being the lowest-level road, starting with highway and ending with pathway. More specific road classifications are detailed in A.3. In impoverished counties, higher-class roads are less common. In contrast, lower-class roads are more necessary and valuable and often lack annotation in publicly available datasets. Therefore, the identification of lower-class roads holds greater significance. The average recognition ability, i.e., the recall metrics for different classes of roads in 2021, is shown in Figure \ref{ablation}, and the results of 2017 are in A.3. From the figure, we can see that when recognizing the high-class roads, OSM has the best performance, and our framework shows a comparable result. However, for the low-class roads (e.g., Ordinary Road and Village Internal Road), which are dominant roads in impoverished counties, our method shows the highest recall, demonstrating the ability of our proposed framework to help study the low-class roads for vulnerable communities.

\subsection{Generated Road Network Dataset}
\begin{figure}
    \centering
    \includegraphics[width=0.9\linewidth]{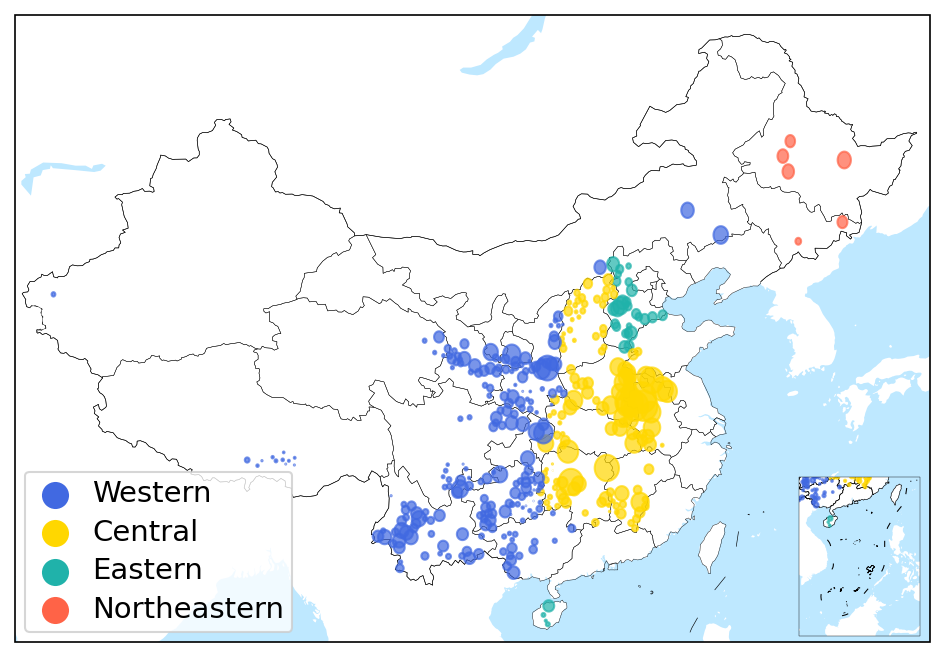}
    \caption{{The spatial distribution of selected impoverished counties. Different colors mark different economic regions and the circle size shows the relative road length in $2021$.}}
    \label{fig:enter-label4}
\end{figure}

We generate the road network dataset covering over $794$,$178$ $km^2$ area in $382$ impoverished counties located in $22$ provinces in China in 2017 and 2021 using the proposed framework. The impoverished counties are selected from the file ``List of 832 National Poverty-Stricken Counties'' \cite{report2} in China. Our proposed road network dataset covers an overall road length of $1.034$ million $km$ and $17.048$ million people in impoverished counties. The counties are assigned to China’s four major economic regions, as shown in Figure \ref{fig:enter-label4}. Most of the impoverished counties are from western and central economic regions, accounting for $90.6\%$ of the selected counties. Besides, we provide the number of images, area, population, Gross Domestic Product (GDP), the added value of the secondary sector of the economy (SSE), and resident saving balance (Balance), road length, road density, and road length per capita in our dataset. Those socioeconomic indicators are collected from the ``China County and City Statistical Yearbook''\footnote{\href{http://www.stats.gov.cn/zs/tjwh/tjkw/tjzl/202302/t20230215_1908004.html}{http://www.stats.gov.cn/zs/tjwh/tjkw/tjzl/202302/
t20230215\_1908004.html}}. The detailed information on the road network dataset and economic indicators of impoverished counties in different economic regions are in B.1.

\section{Socioeconomic Analysis}
In this step, we analyze the scaling law of road networks with respect to population, the correlation between the road network and socioeconomic indicators, and the estimated impact of road networks on the economy. We select GDP, SSE, and balance as the indicators reflecting the regional economic development. Analysis about the variation in road length with respect to population and economic regions are shown in B.2 and B.3, respectively.

\subsection{Scaling Law of Road Network}
The scaling law describes the relationship between socioeconomic factors and infrastructure development as cities grow \cite{bettencourt2013origins}.
It helps in studying and predicting infrastructure correlation with population growth. In this section, we study the scaling law between the extracted road network representing the infrastructure status and population in impoverished counties. The scaling law is defined as
\begin{equation}\label{scaling}
    Y = cX^z,
\end{equation}
where $Y$ is the road infrastructure, i.e., road length in our dataset, $X$ is the population, and $c$ and $z$ are the estimated parameters. 

\begin{figure}[htb]
    \centering
    \includegraphics[width=0.67\linewidth]{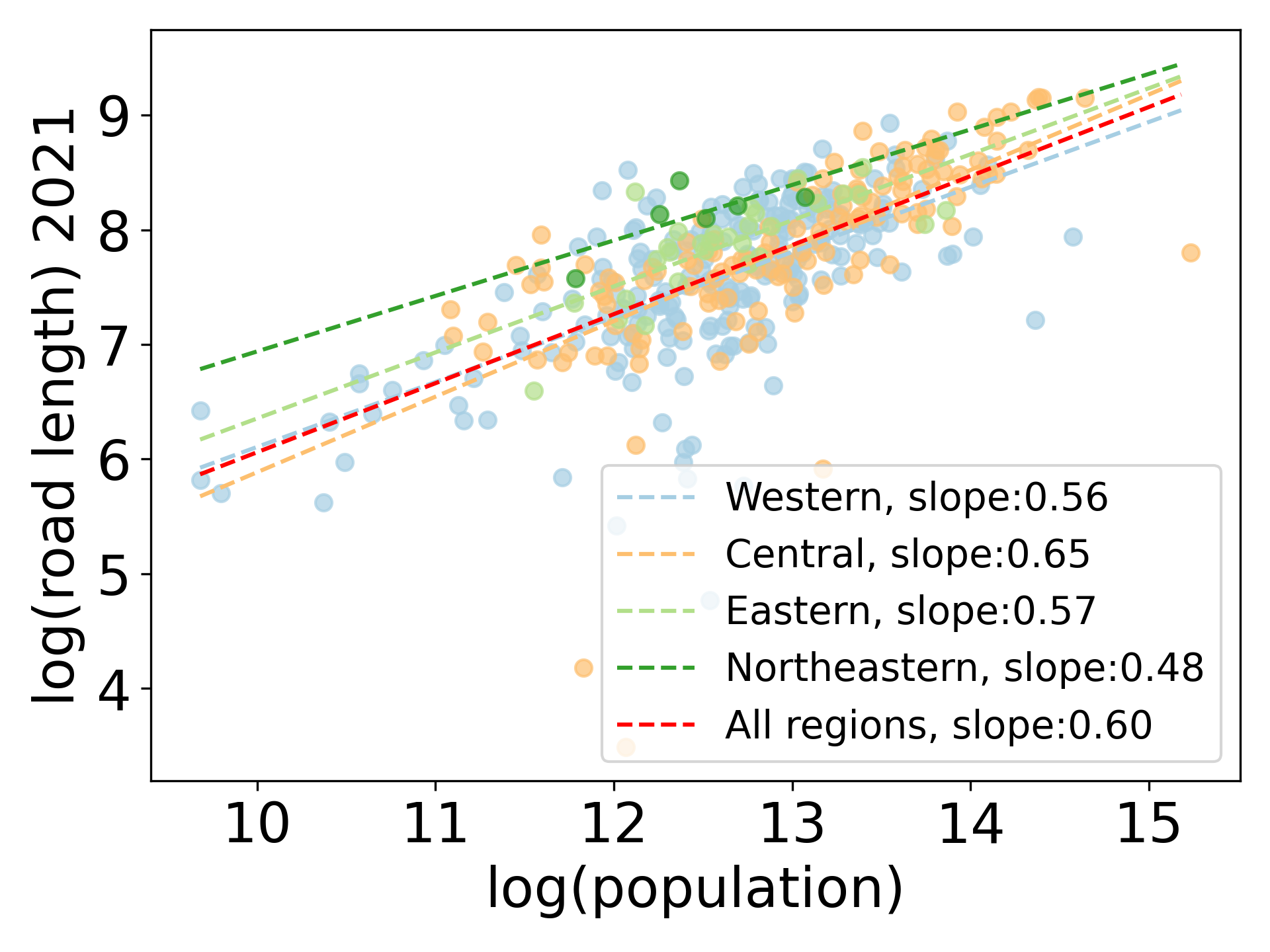}
    \caption{The scaling law between population and road length in different economic regions in $2021$.}
    \label{fig:scaling}
\end{figure}

The fitting curves between the log(population) and log(road length) in $2021$ is shown in Figure \ref{fig:scaling}. Overall, the slope for all counties is $0.6$, showing a sublinear form. The slope is similar to the findings of network length in theoretical computation \cite{bettencourt2013origins}. Then, from the perspectives of different regions, the slopes of the western and central regions in $2021$ show a big gap, demonstrating the uneven development of the road network spatially. The slope for the northeastern region is relatively low, showing that the road network might get saturated. The analysis of year $2017$ is shown in B.4.


\subsection{Correlation Analysis of Road Network}
\begin{figure}[t]
  \centering
  \subfigure[Population]{
    \centering
    \includegraphics[width=0.47\linewidth]
    {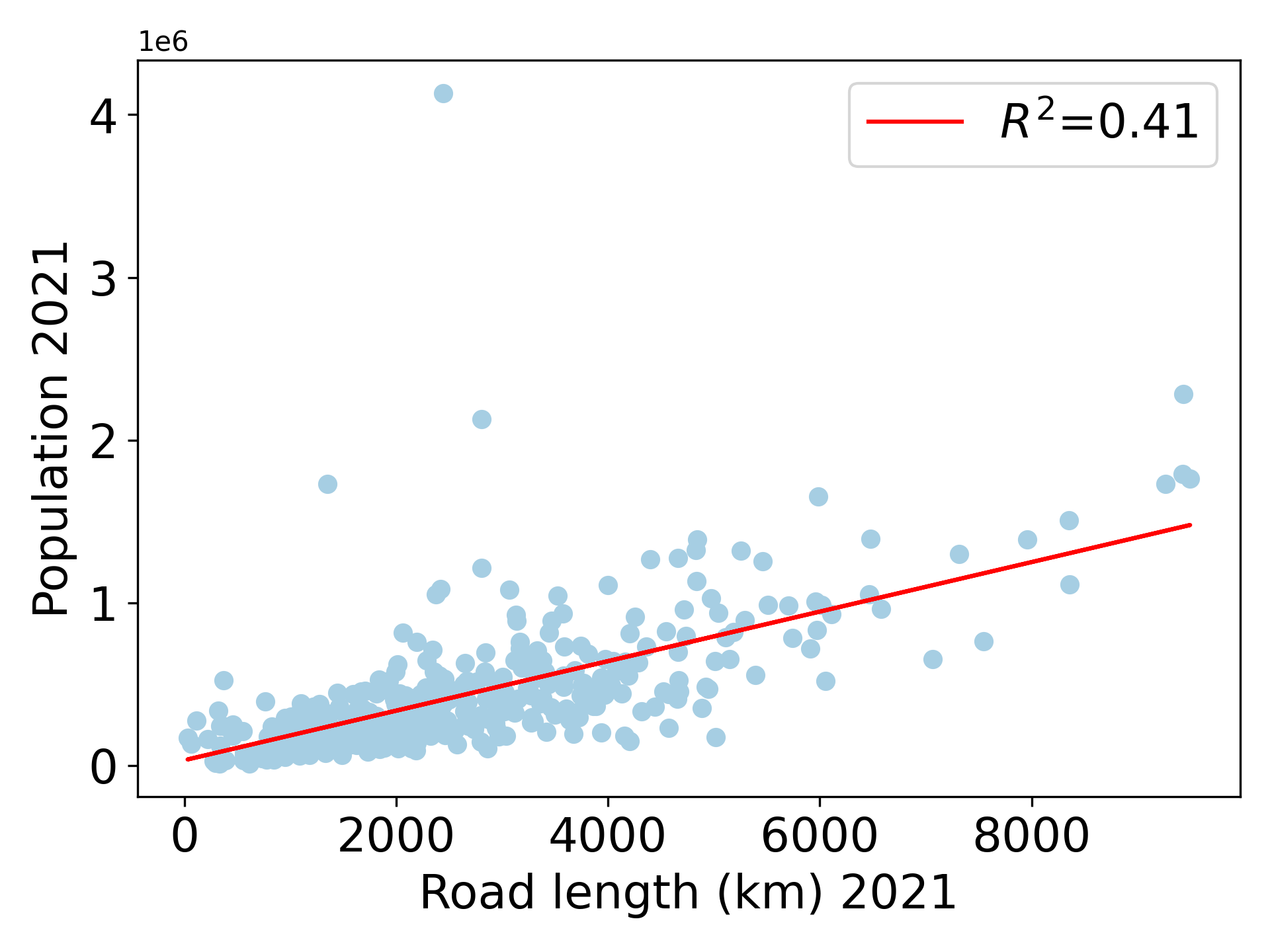}
    }
  \subfigure[GDP]{
    \centering
    \includegraphics[width=0.47\linewidth]{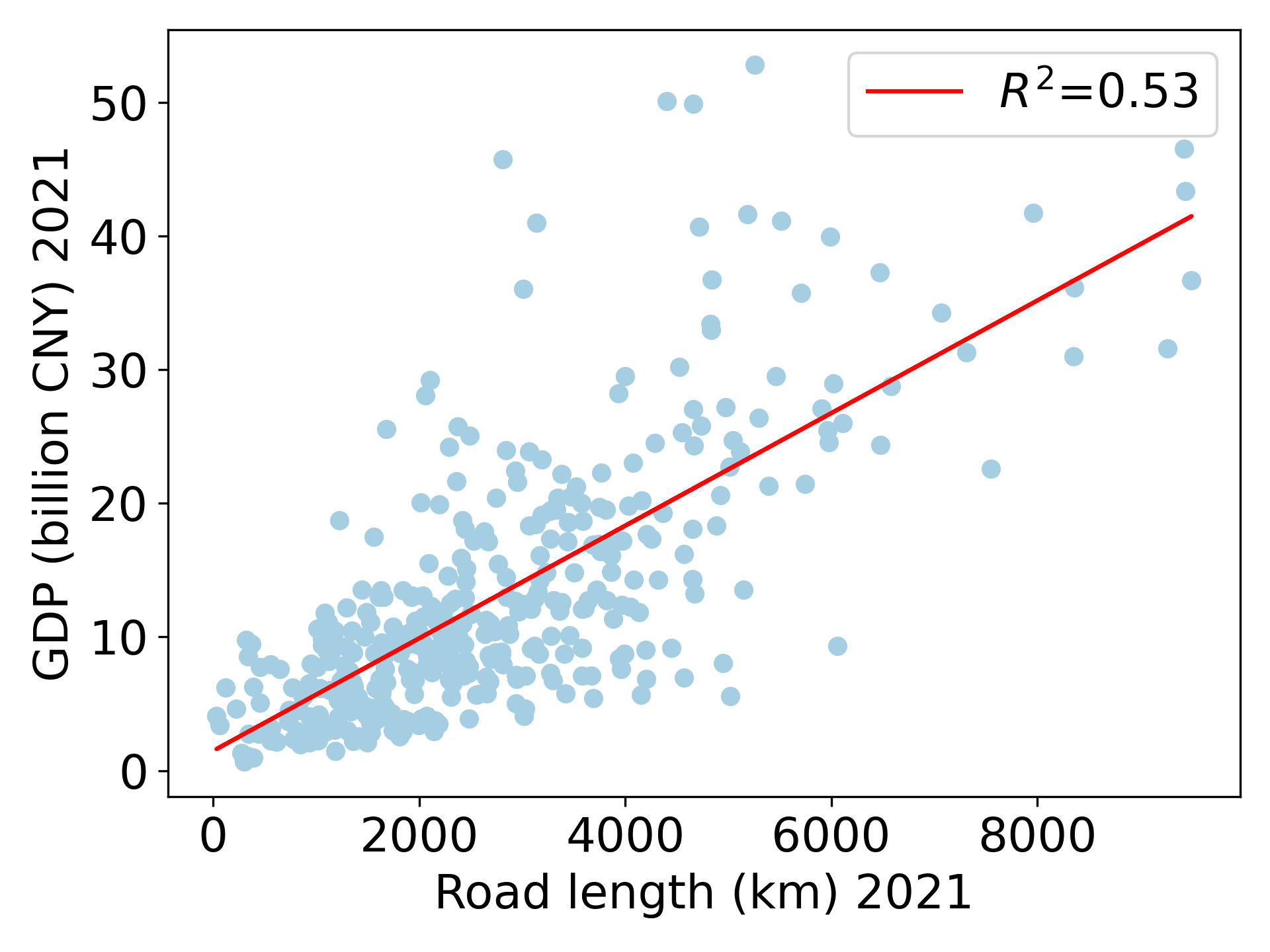}
    }
  \\
    \subfigure[SSE]{
    \centering
    \includegraphics[width=0.47\linewidth]{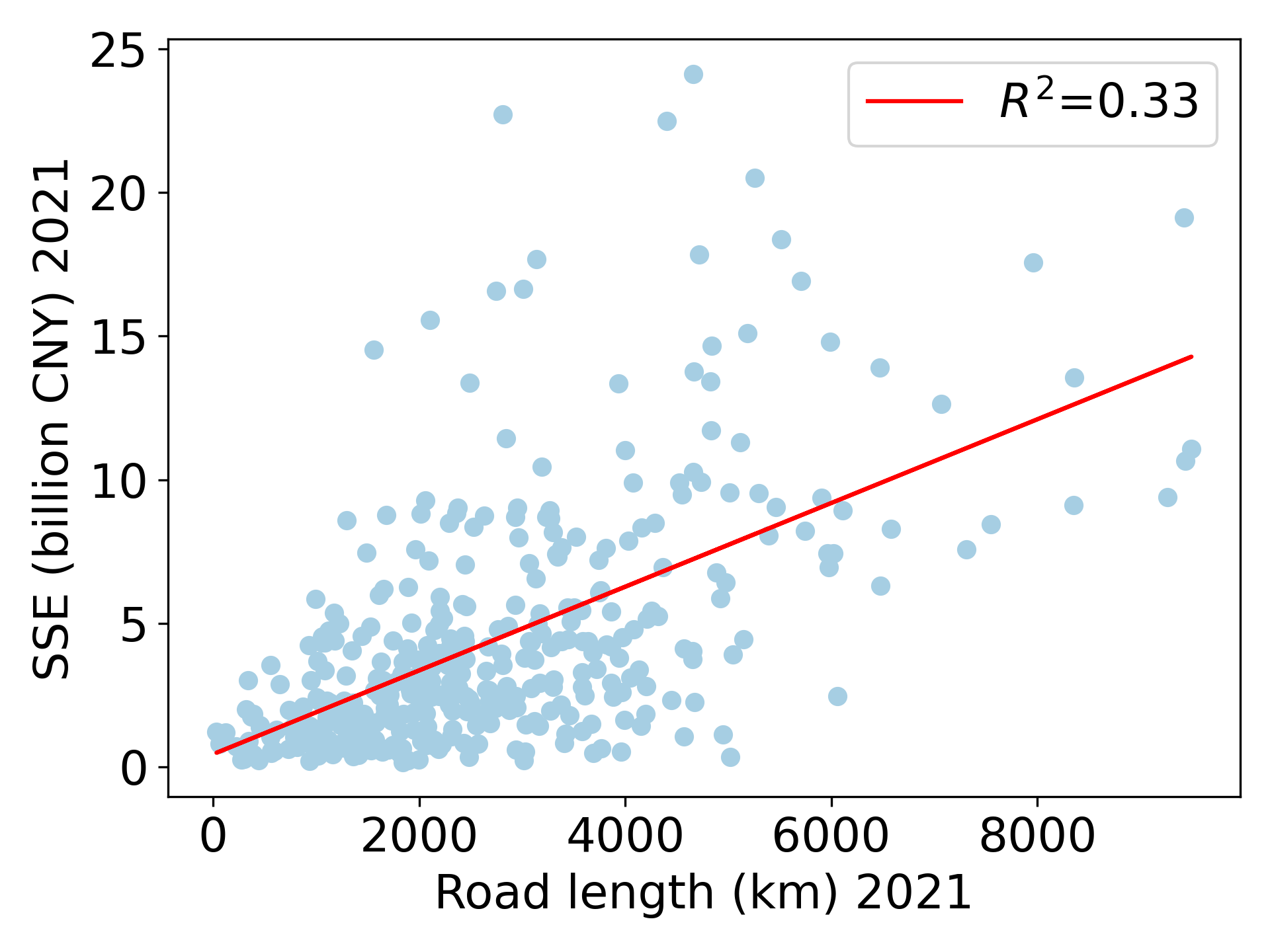}
    }
    \subfigure[Balance]{
    \centering
    \includegraphics[width=0.47\linewidth]{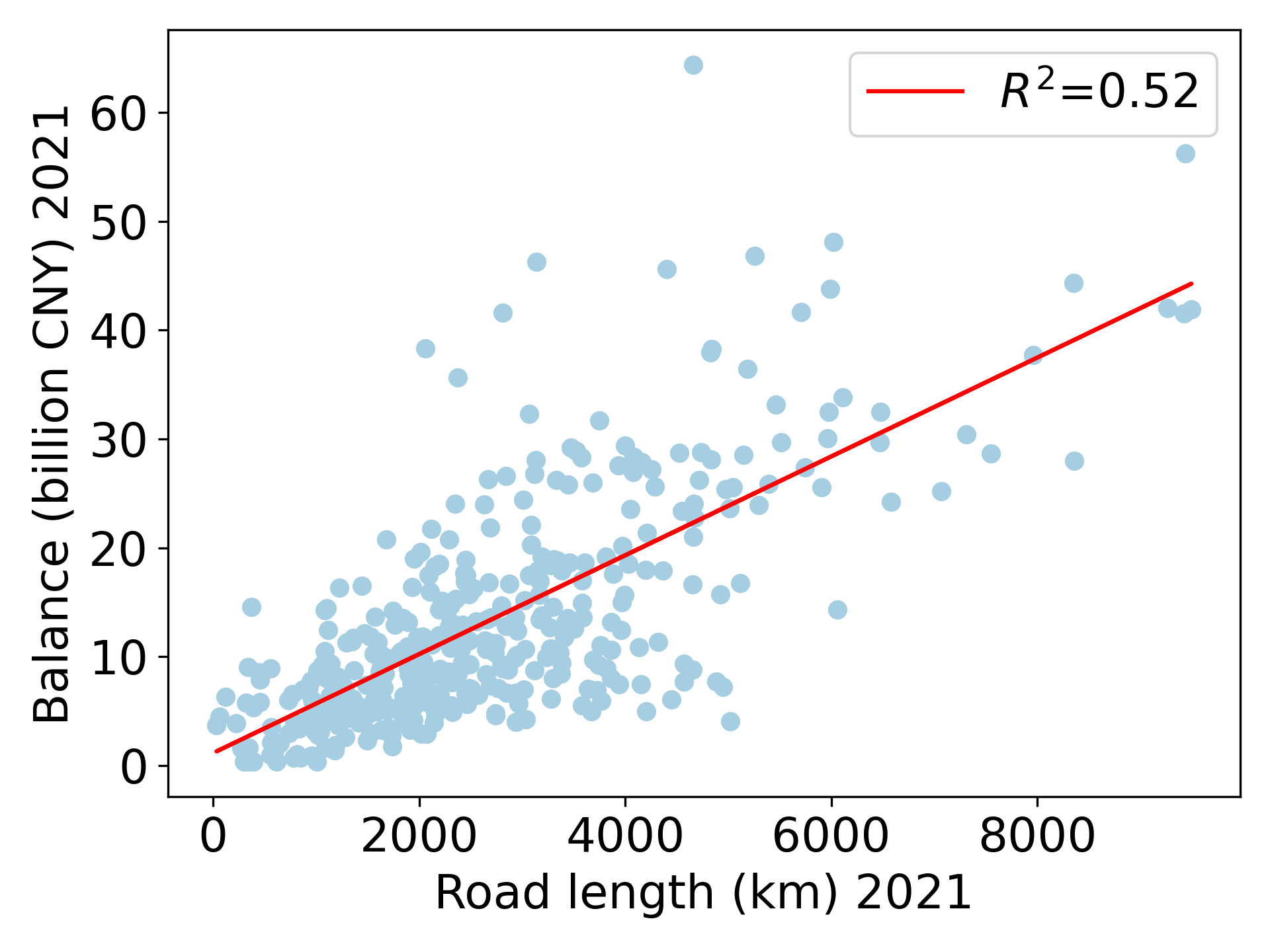}
    }
  \caption{Correlation between the road length and socioeconomic indicators, i.e., population, GDP, SSE, and Balance in impoverished counties in $2021$.}
  \label{fig:correlation}
\end{figure}

The road network is considered the backbone of an area and is related to regional socioeconomic status. In this part, we analyze the correlation between the road network length and regional socioeconomic indicators, i.e., population, GDP, SSE, and balance. The correlation between road length and socioeconomic indicators in 2021 is shown in Figure \ref{fig:correlation}. The correlation results in $2017$ are shown in B.5. 
We can see that the road length and GDP show the highest $R^2$ \cite{lewis2015applied}, confirming that road network length can be used as an indicator for GDP. Second, the resident saving balance shows a similar $R^2$. However, the population and SSE show low $R^2$.
These findings are of importance for subsequent research on vulnerable populations. By observing changes in the road network, a rough estimate of socioeconomic indicators can be obtained, facilitating the formulation and implementation of subsequent policies. 

\subsection{Estimated Impacts of Road Network}
In this step, we study the economic impact of only road networks in impoverished counties with our produced road network dataset, i.e., neglecting other factors such as investment, whether the road network growth from the year $2017$ to $2021$ has a positive impact on regional GDP, SSE, and Balance. Specifically, we apply the Difference-in-difference (DiD) \cite{abadie2005semiparametric} method to study the causal effects of road networks.

We implement the DiD as a fixed effects regression \cite{ratledge2022using}:
\begin{equation}
    Y_{i,t} = \beta (D_i\times T_i) + \gamma_i + \delta_t +\epsilon_{it},
\end{equation}
where $i$ is the subscript for county $i$, and $t$ is time, $Y_{i,t}$ is the socioeconomic indicator of county $i$ in year $t$, $\beta$ is the treatment effects, $(D_i\times T_i)$ is the product of dummy variables $D_i$ (treatment or control group) and $T_i$ (pre- or post-treatment period), $\gamma_i$ is the unit fixed effect, $\delta_t$ is the time fixed effect, and $\epsilon_{it}$ captures the errors.

We divide the counties into control and treatment groups according to the metrics: absolute road length (absolute RL) variation, relative road length (relative RL) variation, and relative road length per capita (relative RPC) variation from year $2017$ to $2021$. The calculations are provided in B.6. For each metric, the counties that belong to the bottom $50\%$ of counties are selected as the control group, and the counties belonging to the top $40\%$ of counties are viewed as the treatment group. In our dataset, 2021 is the post-treatment and 2017 is the pre-treatment period. 

\begin{figure}[t]
    \centering
    \includegraphics[width=0.7\linewidth]{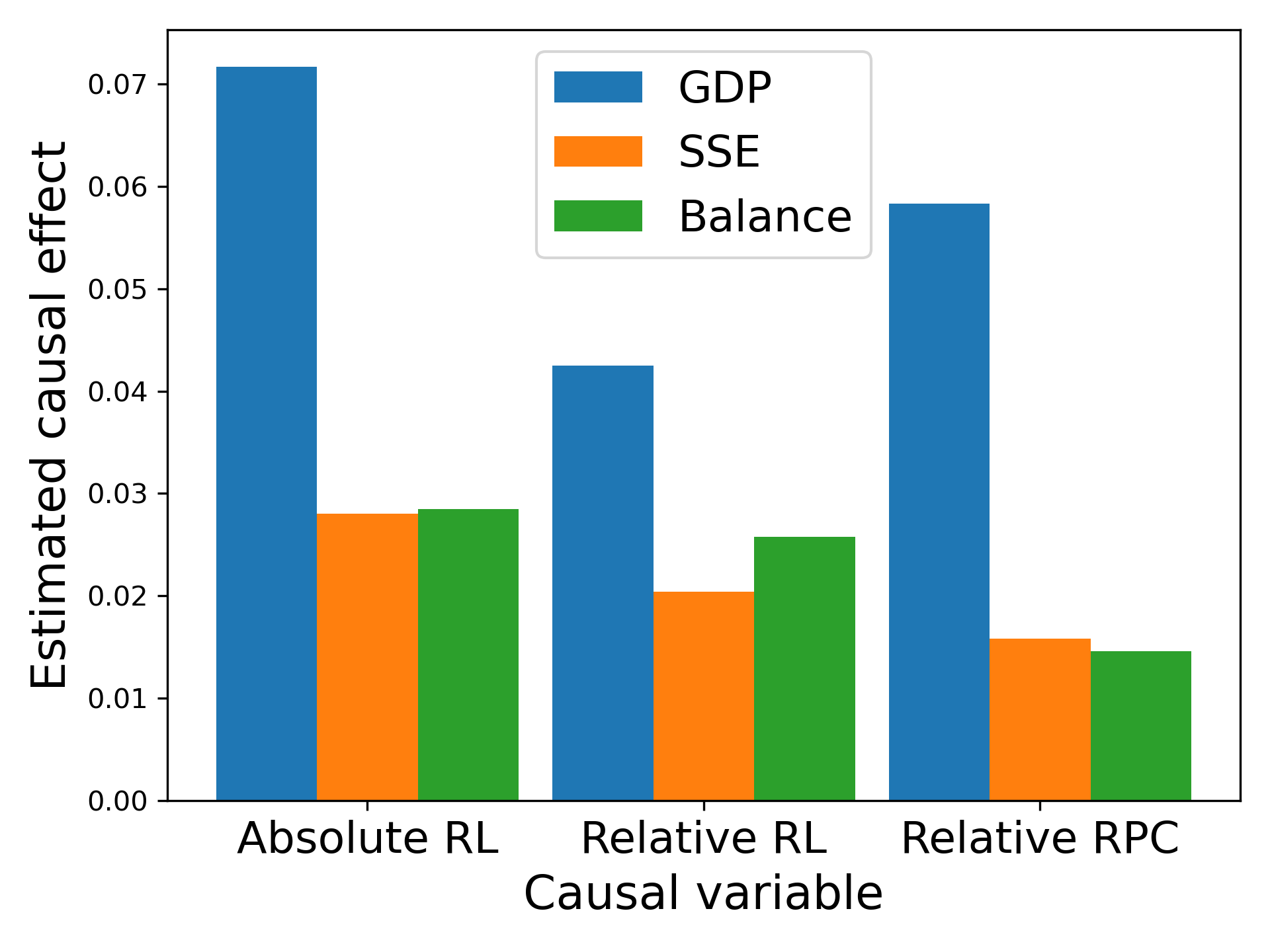}
    \caption{The estimated causal effect of road network construction in impoverished counties.}
    \label{fig:did}
\end{figure}

The estimated causal effects of road networks on GDP, SSE, and balance are shown in Figure \ref{fig:did}. GDP is most affected by road network, and the absolute improvement of road network length shows the highest effect. The relative road network length growth and relative per capita road length also show positive effects. However, many other factors, such as investment, consumption, and inflation \cite{anbao2011comparative,alagic2017analysis}, may influence economic growth and will be analyzed in future work.

\section{Conclusion}
This paper proposes a scalable framework for extracting road networks in impoverished counties from satellite imagery. Extensive experiments on real-world road data validate the effectiveness of the framework. Besides, we generate a comprehensive road network dataset covering $382$ impoverished counties. Performing socioeconomic analysis on our generated road network dataset shows constructing road infrastructure positively correlates with economic development.

Our proposed framework can help with monitoring SDG processes such as poverty eradication or infrastructure accessibility in many impoverished counties where the publicly available road data is not complete. Our work also contributes to the goal of LNOB. By assessing the roads connecting various regions, we provide improved access and opportunities for impoverished areas, thus helping ensure access to education, healthcare, and other essential services.


\section*{Acknowledgements}
This paper is supported in part by the National Natural Science Foundation of China under Grant U22B2057, Grant 62171260, and Grant U21B2036.
\bibliographystyle{named}
\bibliography{ijcai24}

\clearpage
\appendix
\section{Additional Information on Experiments}
\subsection{Validation Dataset Statistics} \label{valData}
To assess the performance of our proposed framework for road network extraction, we randomly select $10$ impoverished counties from those declared out of poverty in recent years in the file ``List of 832 National Poverty-Stricken Counties'' \cite{report2}, whose statistics are shown in Table \ref{tab:val_statis10}. The unit for the average population is ``million people'', the unit for average GDP is ``million Chinese Yuan (CNY)'', and the unit for average area is $km^2$.
\begin{table}[t]
    \centering
    \caption{Statistics of the selected $10$ impoverished counties for road network extraction performance comparison.}
    \begin{tabular}{c|cccc}
    \hline
    Year & Avg. popu &Avg. GDP& Avg. Area & Avg. \#Images\\
    \hline
    2017&0.4218&0.8930&2079&8045\\
    2021&0.4463&1.288&2079&8045\\
    \hline
    \end{tabular}
    \label{tab:val_statis10}
\end{table}

The selected impoverished counties' names are ``Shufu County'', ``Xixiang County'', ``Guanghe County'', ``Danfeng County'', ``Jiangzi County'', ``Honghe County'', ``Libo County'', ``Linquan County'', ``Jingyu County'', and ``Lingqiu County''.

\subsection{Evaluation Metrics} \label{eval-graph}
Graph sampling \cite{biagioni2012inferring,aguilar2021graph} is a commonly used method for comparing two graphs, which simultaneously evaluates the geography and topology of graphs. In our evaluation process, the core idea is to first compute a set of point samples separately on the ground truth road network $G$ and the proposed road network $H$. The point samples are determined by a fixed sampling interval on the road edge, which we set as $0.01$. Then, the selected point samples on $G$ and $H$ are matched one-to-one globally according to a maximum matching distance, which we set as $0.1$ by experiment trials. 

The precision, recall, and F1-score showing the accuracy of the road networks are computed as follows,
\begin{align}
    precision&= \frac{\#matched\ samples}{\#samples\ on\ H},\\ 
    recall &= \frac{\#matched\ samples}{\#samples\ on\ G}, \\
    F1-score &= \frac{2*precision*recall}{precision+recall}.
\end{align}

Then, the road intersection recovery rate (RI) is used for further evaluation. RI reflects the recovery percentage of road intersections in the ground truth road network, and we select the intersection of over three road segments. Besides, Mean Absolute Percentage Error over road length (MRL) and road density (MRD) are adopted for evaluation. MRL and MRD measure the difference between the road length and road density from extracted road network $H$ and ground truth $G$, respectively. The metrics are calculated as
\begin{align}
    MRL&= \frac{1}{N}\sum^{N}_i\frac{|RoadLength(G_i)-RoadLength(H_i)|}{RoadLength(G_i)}, \\
    MRD&= \frac{1}{N}\sum^{N}_i\frac{|RoadDensity(G_i)-RoadDensity(H_i)|}{RoadDensity(G_i)}, \\
    RI\texttt{@}k&= \frac{1}{N}\sum^{N}_i\frac{\#Matched\ intersections\texttt{@}k}{\#Intersections\texttt{@}k\ on\ G_i},
\end{align}
where $N$ denotes the number of selected counties for evaluation, and $RI\texttt{@}k$ denotes the road intersection recovery rate over $k$ road segments($k=3$).

\subsection{Road Classification} \label{roadClassification}
There are following $10$ classes of roads used in the experiments. They are classified based on their width and speed restriction. From the highest to the lowest level, which is the same order used in the experiments, the exact road class names are  ``Highway'', ``National Road'',``Provincial Road'', ``County Road'', ``Township Road'', ``Village Internal Roads'', ``Major Road'', ``Minor Road'', ``Ordinary Road'', and ``Pathway''. 

\begin{figure}[t]
    \centering
    \includegraphics[width=0.9\linewidth]{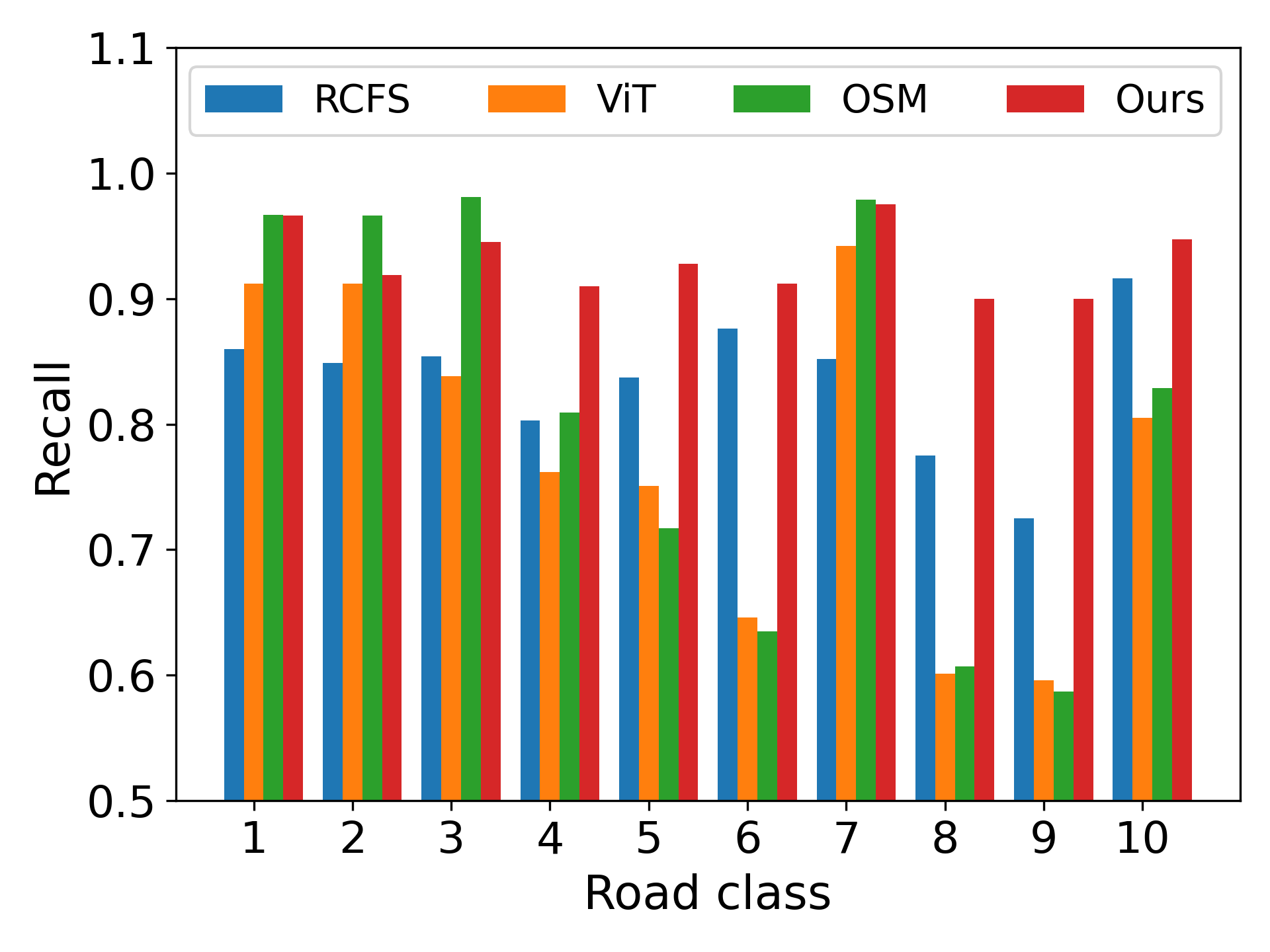}
    \vspace{-10px}
    \caption{The results of recovering different classes of roads in $2017$.}
    \label{ablation2}
\end{figure}

We also show in Figure \ref{ablation2} the recognition results of the above mentioned roads in 2017. We can see that our proposed framework performs well on reconstructing the low-class roads that are important and usually missing in the public dataset.

\section{Dataset Description and Additional Socioeconomic Analysis}
\subsection{Dataset Characteristics} \label{Dataset Characteristics}

\begin{table*}
    \centering
    \begin{tabular}{c|cccccccccccc}
    \hline
    &Regions &\#County  &\#Prov &\#Image &Area &Popu & GDP & SSE & Balance& RL&  RD& RPC\\
    \hline
    \multirow{5}{*}{2017} &Western&214&12&33105&2210&355.8&7.513&2.920&6.363&1878&0.95&6.61\\
                         &Central&132&6&31079&1999&572.0&11.75&4.783&11.50&2537&1.43&5.74\\
                          &Eastern&30&2&20518&1171&362.0&7.447&2.710&8.922&2444&2.68&7.66\\
                    & Northeastern&6&2&81775&3720&283.3&6.601&2.681&5.713&3265&1.08&12.5\\
   \hdashline
                    &All combined&382&22&32180&2079&428.7&8.930&3.543&8.328&2172&1.26&6.48\\
    \hline
    \multirow{5}{*}{2021} &Western&214&12&33105&2210&370.1&10.75&3.576&9.776&2340&1.17&8.06\\
                        &Central&132&6&31079&1999&597.2&17.31&6.242&19.04&3236&1.85&7.05\\
                        &Eastern&30& 2&20518&1171&360.0&9.887&2.696&15.64&2842&3.17&9.05\\
                    &Northeastern&6&2&81775&3720&275.0&6.546&1.322&9.695&3484&1.19&13.8\\
    \hdashline
                    &All combined&382&22&32180&2079&446.3&12.88&4.393&13.44&2707&1.56&7.88\\
    \hline
    \end{tabular}
    \caption{{{Dataset statistics of the selected impoverished counties in different economic regions in 2017 and 2021.}}}
    \label{tab:val_statis382}
\end{table*}


We leverage the satellite image to generate a road network dataset targeting impoverished areas to support SDGs. And $382$ impoverished counties are selected from those declared out of poverty in recent years in the file ``List of 832 National Poverty-Stricken Counties'' \cite{report2} in China. The $382$ impoverished counties are assigned to China’s four major economic regions, which are determined by the China State Council according to the socioeconomic development of different regions. Those four regions are the western, central, eastern, and northeastern economic regions. 

We select the Gross Domestic Product (GDP), added value of the secondary sector of the economy (SSE), and resident saving balance (Balance) as the indicators reflecting the regional socioeconomic development. These indicators are collected from the ``China County and City Statistical Yearbook\footnote{\href{http://www.stats.gov.cn/zs/tjwh/tjkw/tjzl/202302/t20230215_1908004.html}{http://www.stats.gov.cn/zs/tjwh/tjkw/tjzl/202302/
t20230215\_1908004.html}}'' published by China Statistics Press in 2017 and 2021. The detailed socioeconomic status of the selected counties in different economic regions are presented in Table \ref{tab:val_statis382}. Altogether, we show the number of counties (\#County), number of provinces (\#Prov), average number of images (\#Image), average area ($km^2$), average population (popu, thousand people), average GDP (billion CNY), average added value of the secondary sector of the economy (SSE, billion CNY), average resident saving balance (Balance, billion CNY), average road length (RL, $km$), average road density (RD, $km/km^2$), and average road length per capita (RPC, $m$). Most of the counties reside in the western and central economic regions, because China's economic development is higher in eastern provinces than other regions and most counties declaring out of poverty are located in the western and central regions. 
From the perspective of the road networks, the western region still has the least road length and road density, and the eastern region almost has twice the road density, which shows the road accessibility inequality among different economic regions. 

\subsection{Road Network Features with County Size} \label{countysize}
\begin{figure}[t]
  \centering
  \subfigure[Road length, 2017]{
    \centering
    \includegraphics[width=0.46\linewidth]{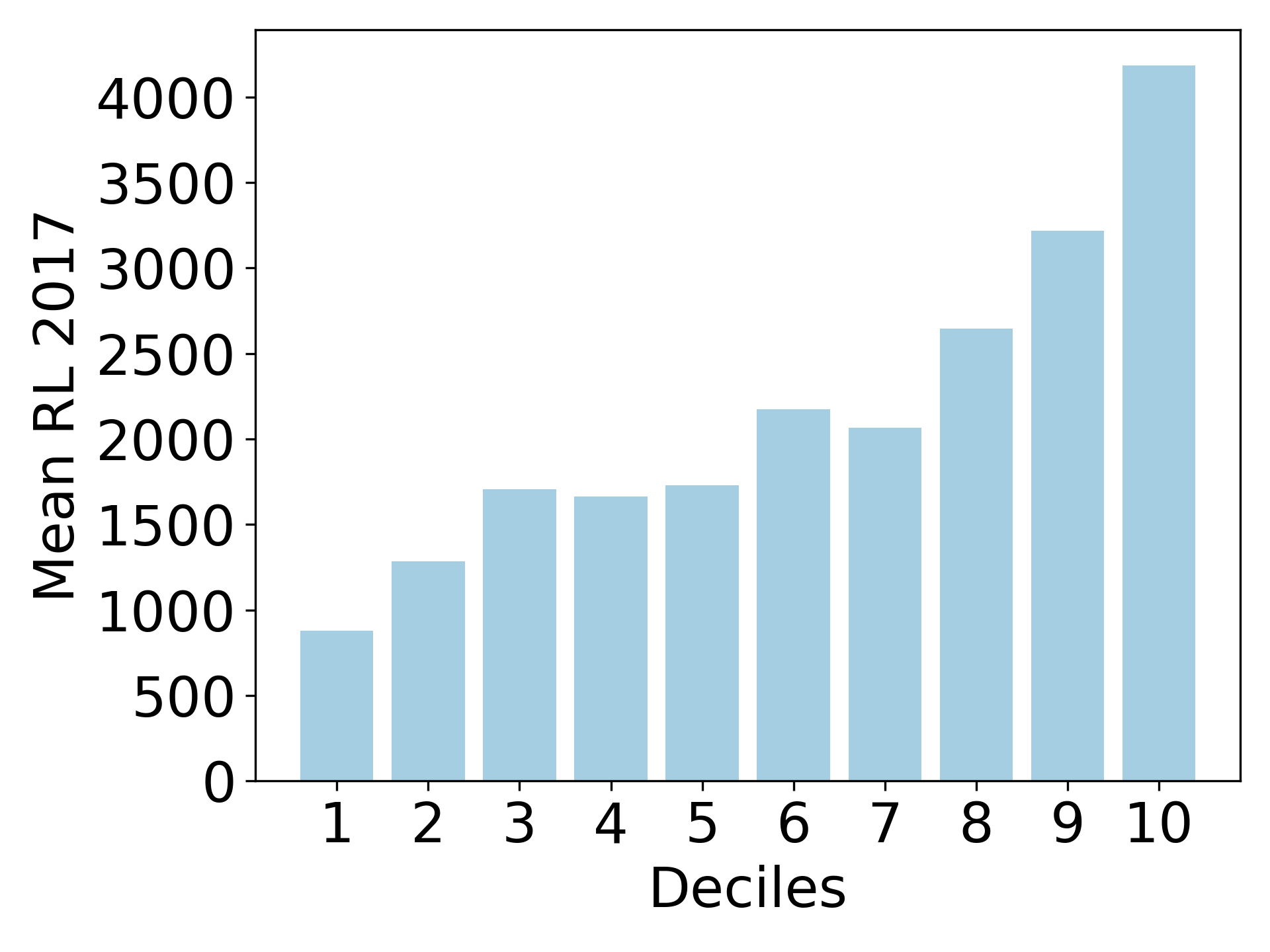}
    \label{fig:sub1}
  }
  \subfigure[Road length per capita, 2017]{
    \centering
    \includegraphics[width=0.46\linewidth]{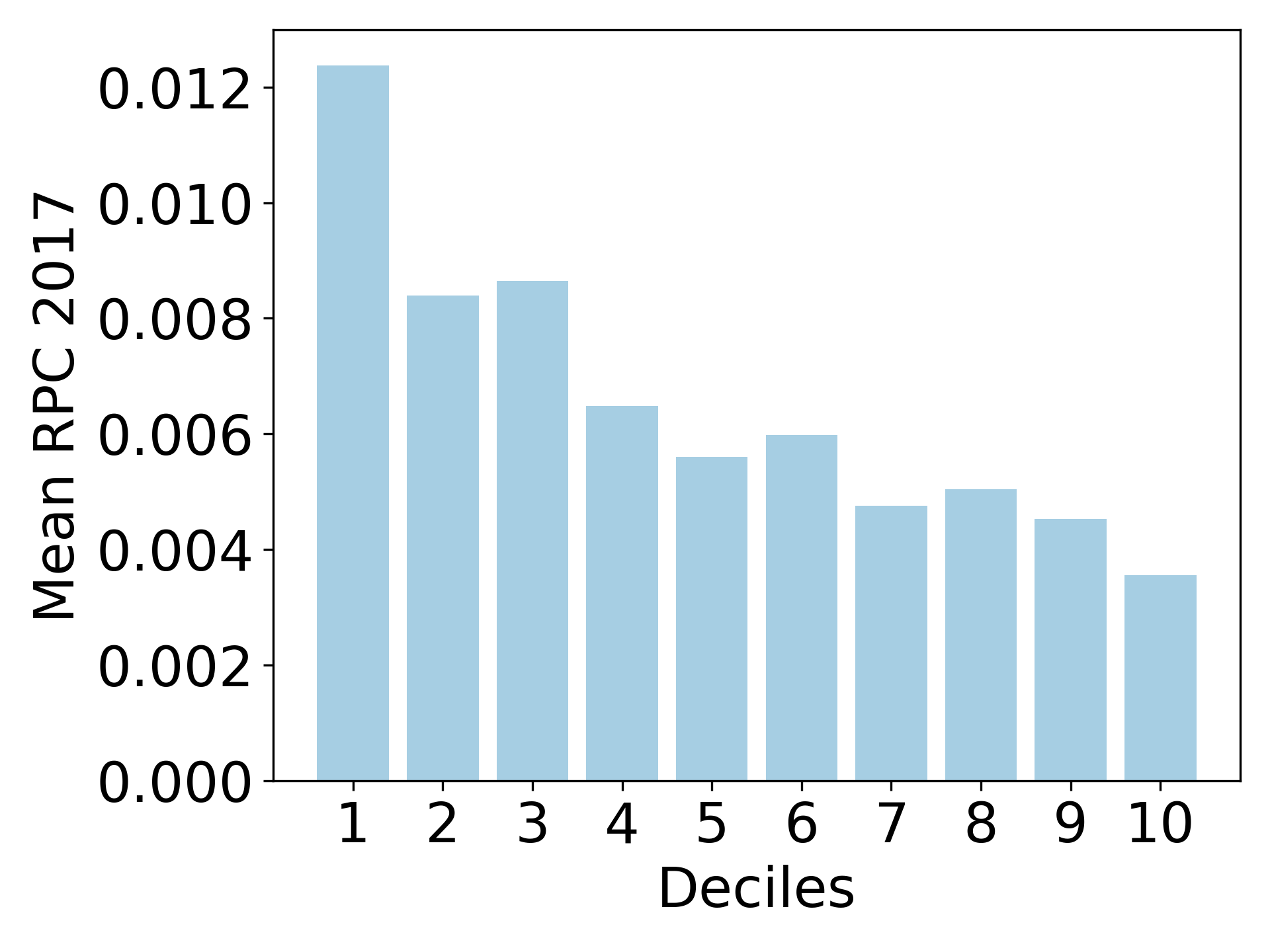}
    \label{fig:sub2}
}
  \\
    \subfigure[Road length, 2021]{
    \centering
    \includegraphics[width=0.46\linewidth]{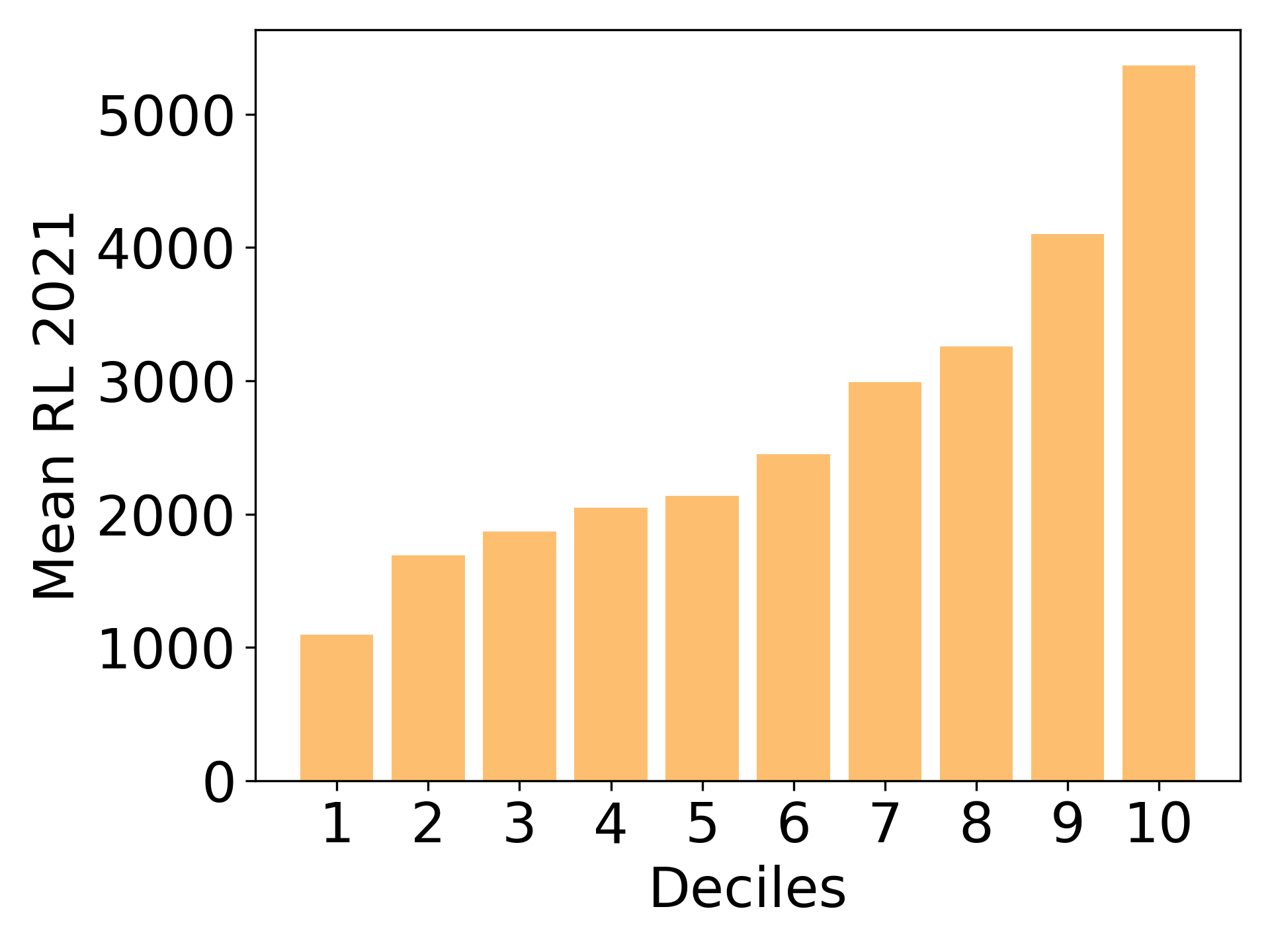}
    \label{fig:sub3}
}
    \subfigure[Road length per capita, 2021]{
    \centering
    \includegraphics[width=0.46\linewidth]{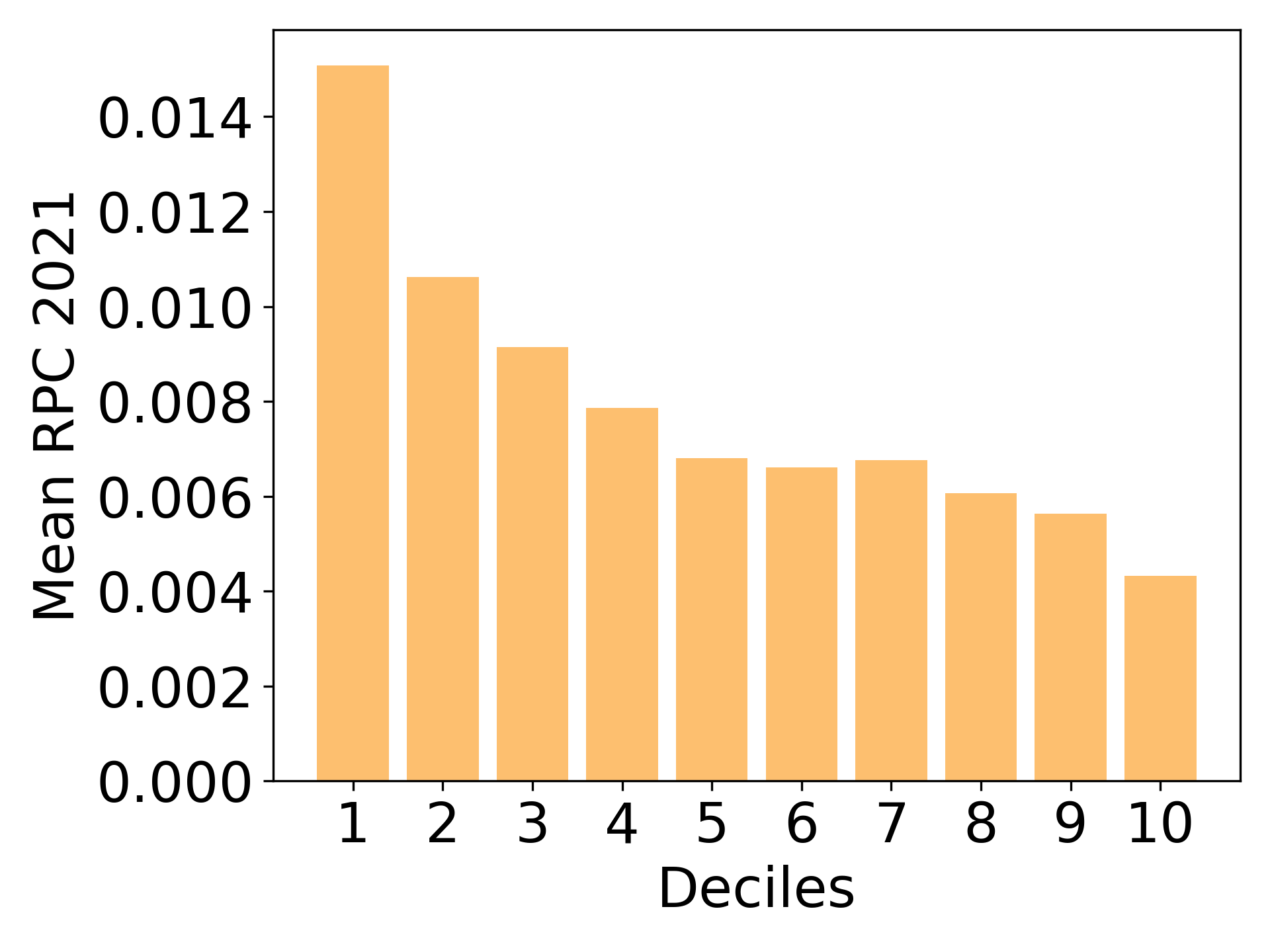}
    \label{fig:sub3}
}
  \vspace{-10px}
  \caption{The comparison of road network length and per capita road length in counties from different population groups.}
  \label{fig:deciles}
  \vspace{-10px}
\end{figure}

In this section, we compare the road network length and per capita road length in our generated dataset across counties with different population size. Specifically, we group the counties into 10 deciles (that means decile 10 represents the counties with the top 10\% population and decile 1 includes the counties with the least 10\% population) and compute the mean road length (RL) and road length per capita (RPC). The results in different deciles in 2017 and 2021 respectively are presented in Figure \ref{fig:deciles}. The results show that counties with more population have more road length, but the expansion of road network is slower than the growth of population, which is consistent with the findings in US metropolitan areas \cite{levinson2012network}.

\subsection{Inequality of Road Network Expansion} 

\begin{figure}[t]
  \centering
  \subfigure[Road length.]{
    \centering
    \includegraphics[width=0.47\linewidth]{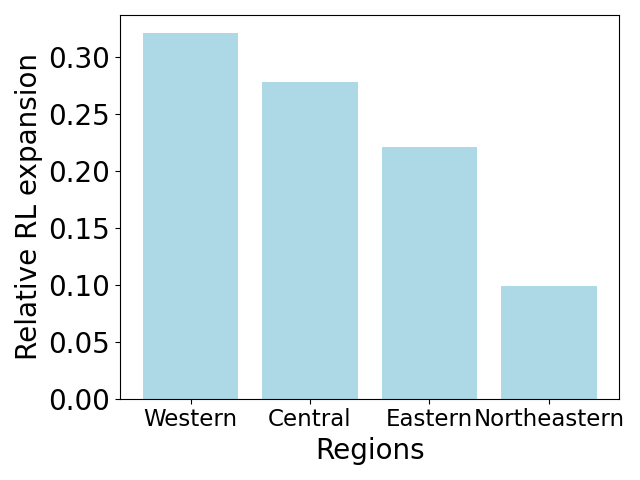}
    \label{fig:sub1}}
  \subfigure[Road length per capita.]{
    \centering
    \includegraphics[width=0.47\linewidth]{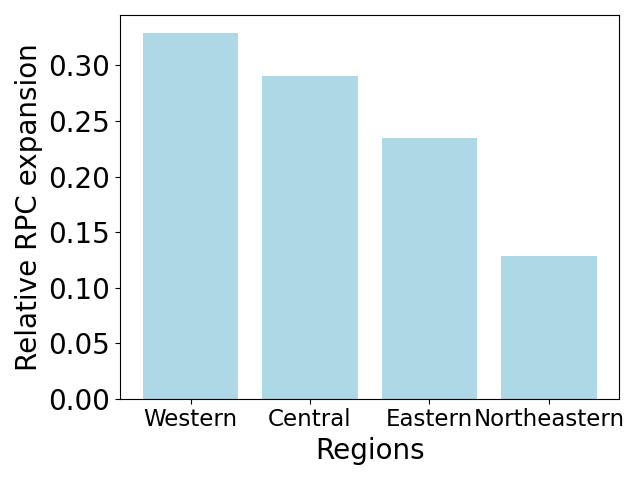}
    \label{fig:sub2}}
  \caption{The expansion of roads in impoverished counties from different groups.}
  \label{fig:macro-comp2}
\end{figure}

The economic development of the four economic regions in China is different, with the eastern region being the most developed and the western region the least developed. Therefore, the difference in road network variation in those regions can help understand the improvement of the local economy. We present mean values of the relative road length (RL) growth and relative road length per capita (RPC) growth from 2017 to 2021 in impoverished counties from four economic regions in Figure \ref{fig:macro-comp2}. We can see that the road length growth in the western region is the highest, which might be due to the efforts of road construction to alleviate poverty and boost the economy in the western region. Next, impoverished counties in the central region experience the second fastest expansion in road infrastructure. 

\subsection{Scaling Law of Road Network in Year 2017}
\begin{figure}[t]
    \centering
    \includegraphics[width=0.65\linewidth]{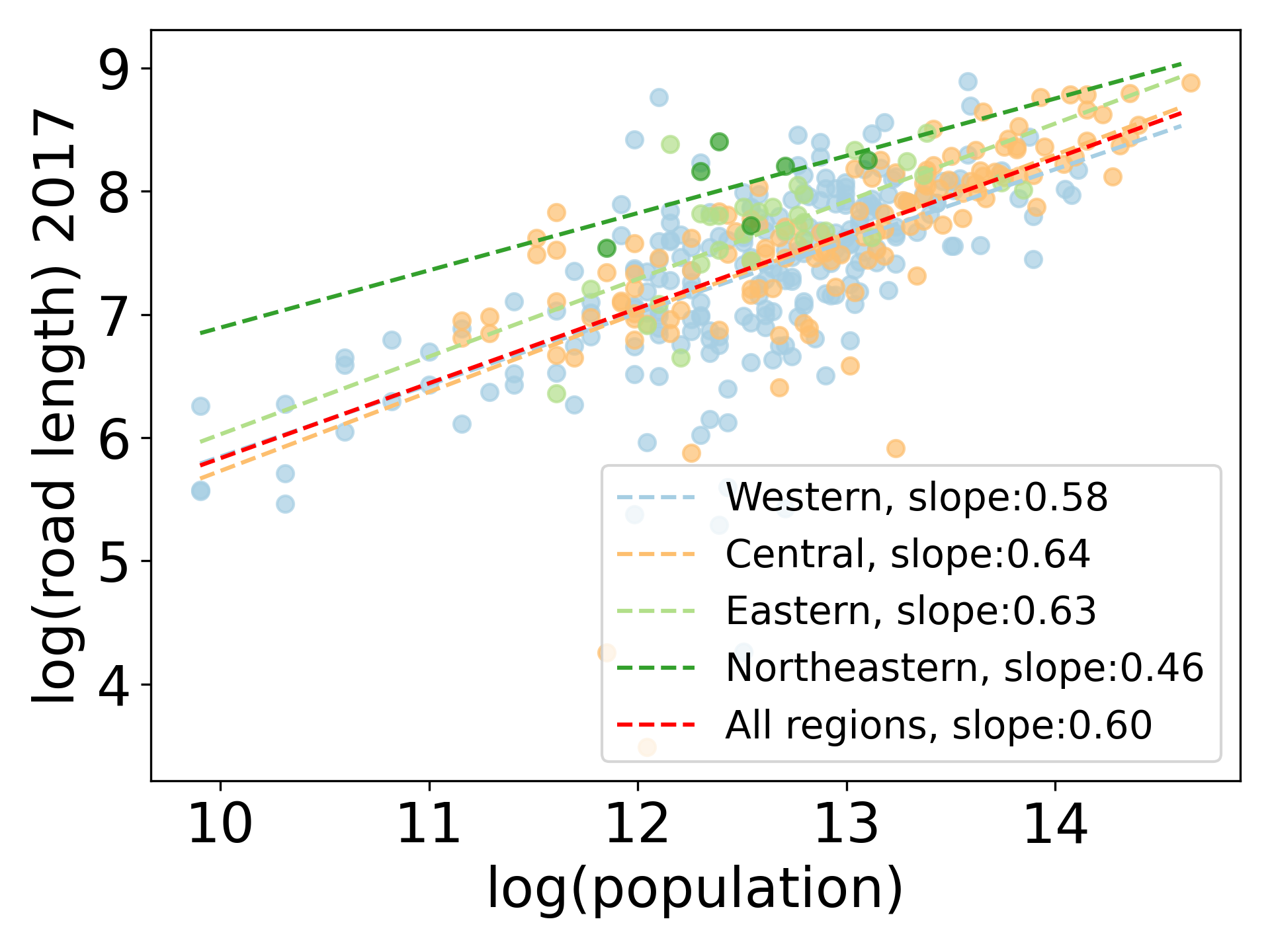}
    \vspace{-10px}
    \caption{The scaling law between population and road length in different economic regions in $2017$.}
    \label{scaling2017}
\end{figure}

The scaling law analysis between the log(population) and log(road length) in 2017 is shown in Figure \ref{scaling2017}. The slopes of the western and central regions in $2017$ demonstrate the uneven development of the road network spatially.

\subsection{Correlation Analysis in Year 2017} \label{corr2017}
\begin{figure}[t]
  \centering
  \subfigure[Population]{
    \centering
    \includegraphics[width=0.46\linewidth]{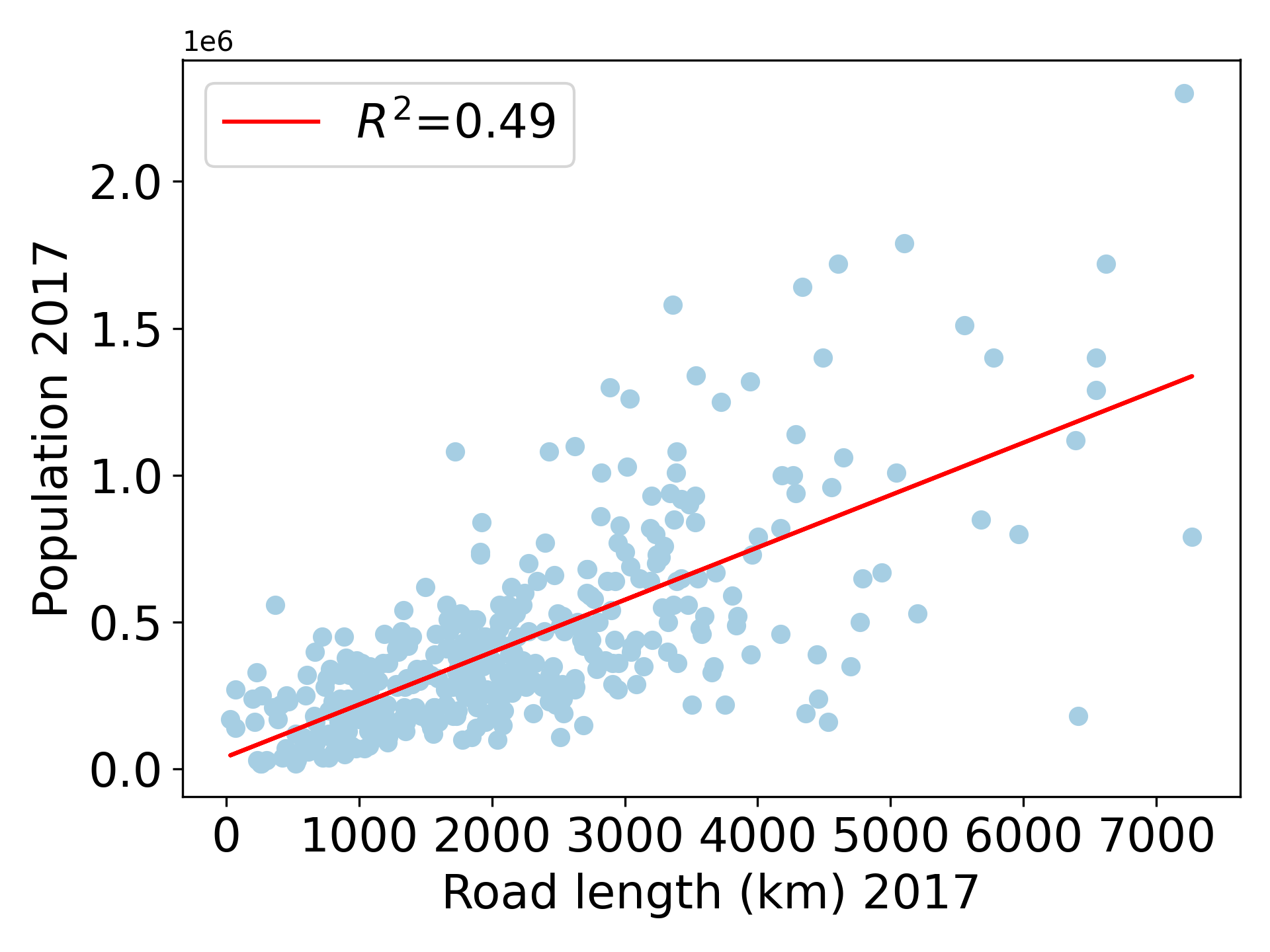}
    \label{fig:sub1}}
  \subfigure[GDP]{
    \centering
    \includegraphics[width=0.46\linewidth]{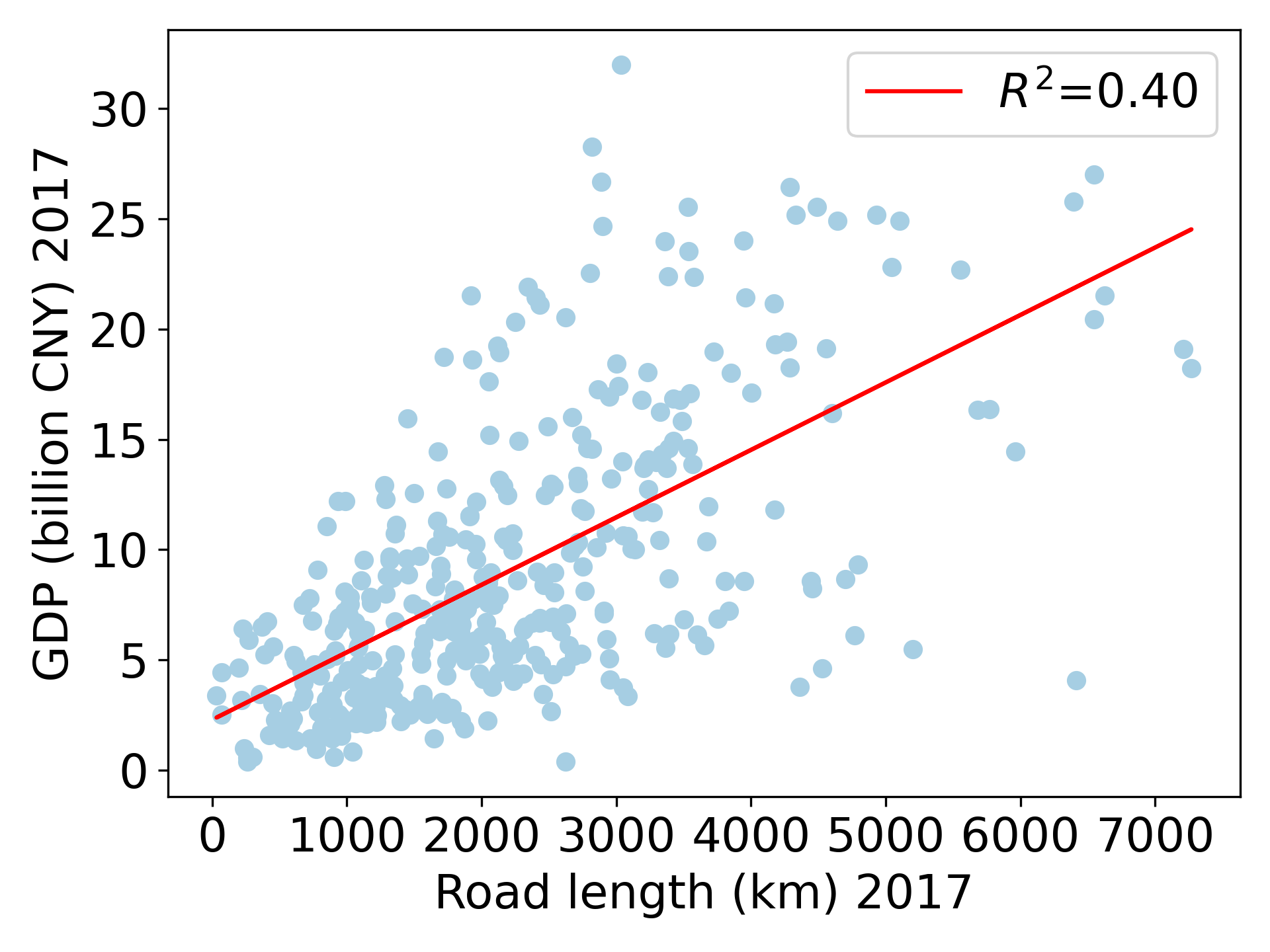}
}
  \\
    \subfigure[SSE]{
    \centering
    \includegraphics[width=0.46\linewidth]{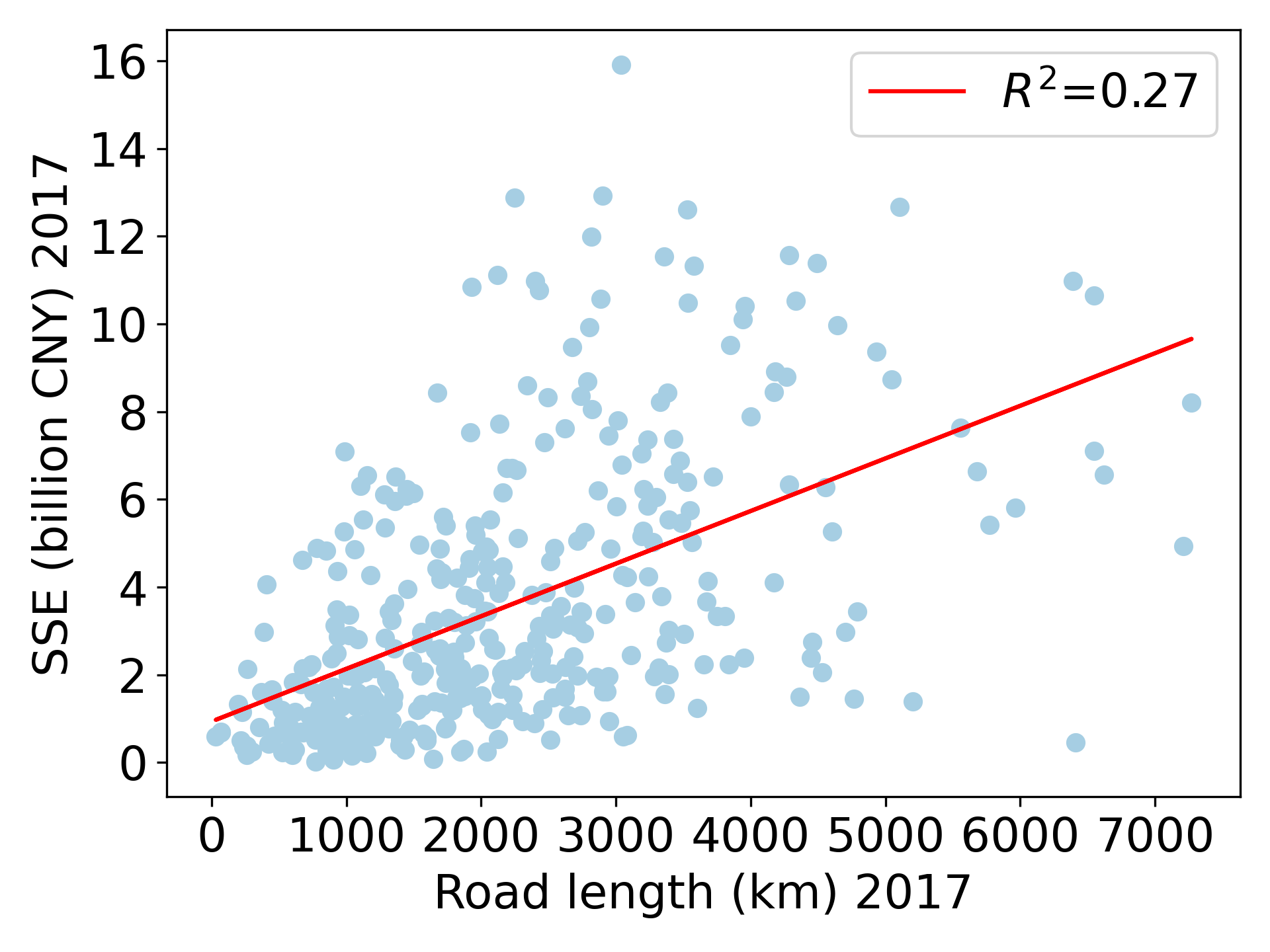}
}
    \subfigure[Balance]{
    \centering
    \includegraphics[width=0.46\linewidth]{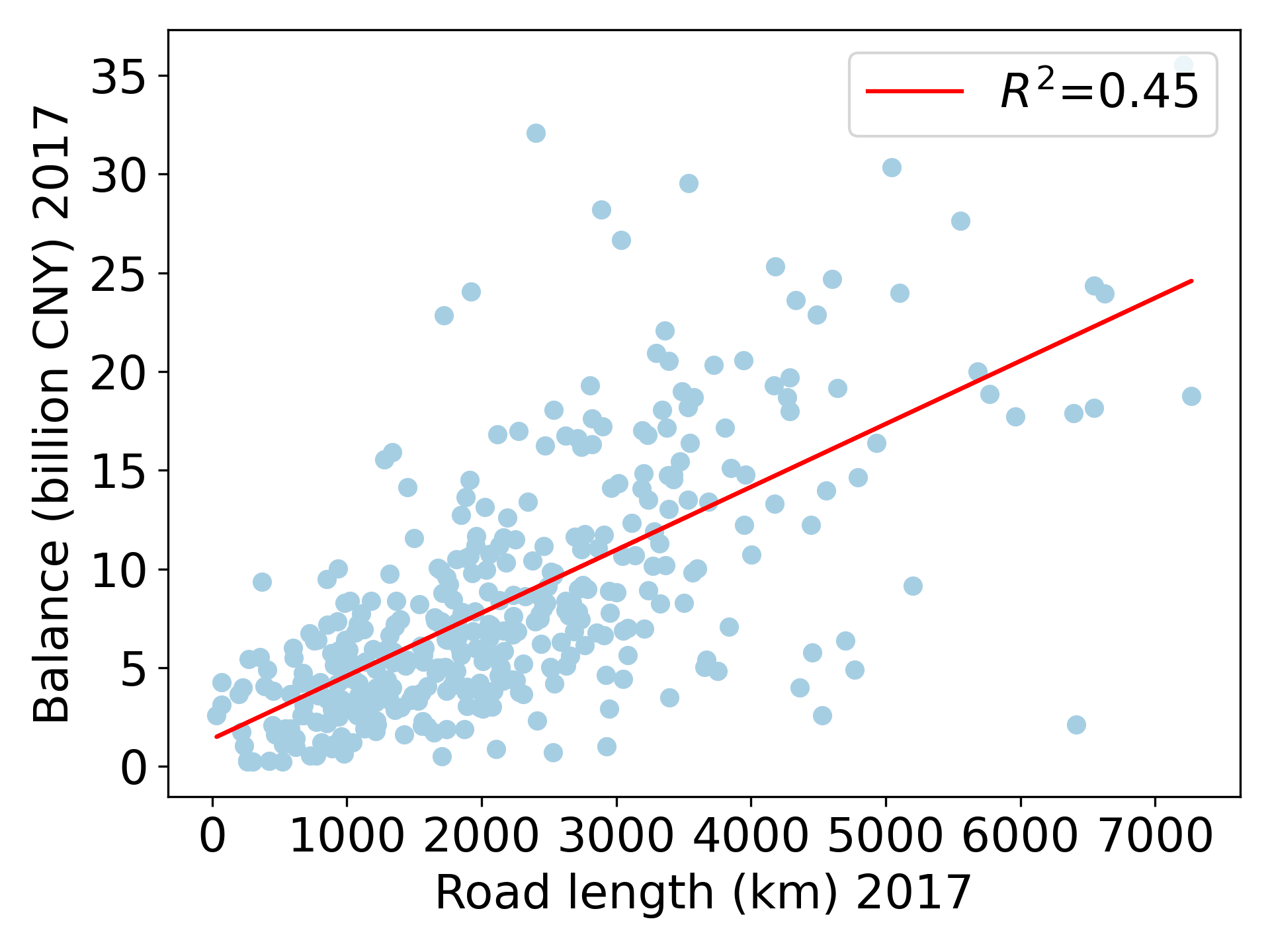}
    }
  \vspace{-10px}
  \caption{Correlation between the road length and socioeconomic indicators, i.e., population, GDP, SSE, and Balance in impoverished counties in 2017.}
  \label{fig:correlation2}
  \vspace{-10px}
\end{figure}

The correlation analysis of road network length and population, GDP, SSE, and balance in our generated datatset in 2017 is shown in Figure \ref{fig:correlation2}. We can see the population, GDP, and balance show higher $R^2$ than SSE.

\subsection{Division of Control and Treatment Groups in Causal Analysis of Road Network} \label{CausalCal}
We divide the counties into control and treatment groups according to the metrics: absolute road length variation (ARL), relative road length variation (RRL), and relative per capita road length variation (RRPC). The calculation formulas are as follow: 
\begin{gather}
    ARL = RL_{\mathrm{t_2}} -RL_{\mathrm{t_1}},\\
    RRL = (RL_{\mathrm{t_2}} -RL_{\mathrm{t_1}})/RL_{\mathrm{t_2}},\\
    RRPC = \frac{(RL_{\mathrm{t_2}}/P_{\mathrm{t_2}} -RL_{\mathrm{t_1}}/P_{\mathrm{t_1}})}{(R_{\mathrm{t_2}}/P_{\mathrm{t_2}})},
\end{gather}
where $RL_{\mathrm{t_2}}$ and $RL_{\mathrm{t_1}}$ are the road length in year $\mathrm{t_2}$ and $\mathrm{t_1}$, and $P$ represents the county's population.



\end{document}